%% file: Main.tex
\documentclass{article}

\usepackage[english]{babel}

\usepackage[a4paper,top=2cm,bottom=2cm,left=3cm,right=3cm,marginparwidth=1.75cm]{geometry}

\usepackage{amssymb}
\usepackage{siunitx}
\PassOptionsToPackage{hyphens}{url}\usepackage{hyperref}
\usepackage[utf8]{inputenc}
\usepackage[right]{lineno}
\usepackage{csquotes}
\usepackage{booktabs}
\usepackage{longtable}
\usepackage{multirow}
\usepackage[table,xcdraw]{xcolor}
\usepackage{lscape}
\usepackage{longtable}
\usepackage{array}
\usepackage{ragged2e}
\usepackage{amsmath}
\usepackage{amssymb}
\usepackage{adjustbox}
\usepackage{array}
\usepackage{url}
\usepackage{titlesec}
\usepackage{authblk}
\usepackage{xcolor} 
\renewcommand{\thetable}{\Roman{table}}
\usepackage{graphicx}
\usepackage{amsmath}
\usepackage{geometry}
\usepackage{cleveref}
\titleformat{\subsection}
  {\mdseries\itshape\large} 
  {\thesubsection}{1em}{} 

\usepackage[english]{babel}
\usepackage[style=authoryear,backend=biber,natbib=true,maxcitenames=2,uniquelist=false]{biblatex}
\DeclareNameAlias{sortname}{family-given}
\DeclareNameAlias{default}{family-given}

\renewbibmacro{in:}{}
\DeclareFieldFormat[article]{title}{\mkbibquote{#1}\addcomma}
\DeclareFieldFormat[book]{title}{\mkbibemph{#1}\addcomma}
\DeclareFieldFormat[bookinbook]{title}{\mkbibemph{#1}\addcomma}
\DeclareFieldFormat[inbook]{title}{\mkbibquote{#1}\addcomma}
\DeclareFieldFormat[incollection]{title}{\mkbibquote{#1}\addcomma}
\DeclareFieldFormat[inproceedings]{title}{\mkbibquote{#1}\addcomma}
\DeclareFieldFormat[manual]{title}{\mkbibemph{#1}\addcomma}
\DeclareFieldFormat[misc]{title}{\mkbibemph{#1}\addcomma}
\DeclareFieldFormat[thesis]{title}{\mkbibemph{#1}\addcomma}
\DeclareFieldFormat[unpublished]{title}{\mkbibquote{#1}\addcomma}
\DeclareFieldFormat[patent]{title}{\mkbibemph{#1}\addcomma}
\DeclareFieldFormat[report]{title}{\mkbibemph{#1}\addcomma}
\DeclareFieldFormat[online]{title}{\mkbibquote{#1}\addcomma}
\DeclareFieldFormat[software]{title}{\mkbibemph{#1}\addcomma}
\DeclareFieldFormat[booklet]{title}{\mkbibemph{#1}\addcomma}
\DeclareFieldFormat[periodical]{title}{\mkbibemph{#1}\addcomma}
\DeclareFieldFormat[standard]{title}{\mkbibemph{#1}\addcomma}

\DeclareFieldFormat[article]{journaltitle}{\iffieldundef{shortjournal}{\mkbibemph{#1}\addcomma}{\mkbibemph{\printfield{shortjournal}}\addcomma}}
\DeclareFieldFormat{volume}{\bibstring{volume}~#1}
\DeclareFieldFormat{number}{\bibstring{number}~#1}

\DefineBibliographyStrings{english}{
  volume = {Vol.},
  number = {No.}
}

\renewbibmacro*{volume+number+eid}{%
  \printfield{volume}%
  \setunit*{\addspace}%
  \printfield{number}%
  \setunit{\addcomma\space}%
  \printfield{eid}}

\renewbibmacro*{journal+issuetitle}{%
  \usebibmacro{journal}%
  \setunit*{\addcomma\space}%
  \usebibmacro{volume+number+eid}%
  \setunit{\addcomma\space}%
  \usebibmacro{issue+date}}

\renewbibmacro*{publisher+location+date}{%
  \printlist{publisher}%
  \iflistundef{location}
    {\setunit*{\addcomma\space}}
    {\setunit*{\addcolon\space}}%
  \printlist{location}%
  \setunit*{\addcomma\space}%
  \usebibmacro{date}}

\DeclareCiteCommand{\cite}[\mkbibparens]
  {\usebibmacro{prenote}}
  {\usebibmacro{citeindex}%
   \usebibmacro{cite}}
  {\multicitedelim}
  {\usebibmacro{postnote}}

\renewbibmacro*{cite:labelyear+extrayear}{%
  \iffieldundef{labelyear}
    {}
    {\printtext[bibhyperref]{%
       \printfield{labelyear}%
       \printfield{extrayear}}}}

\renewbibmacro*{cite:labeldate+extradate}{%
  \iffieldundef{labelyear}
    {}
    {\printtext[bibhyperref]{%
       \printfield{labelyear}%
       \printfield{extradate}}}}

\AtEveryBibitem{
  \clearfield{month}
  \clearfield{day}
  \ifentrytype{book}{
    \clearlist{location}
  }{}
}

\DefineBibliographyStrings{english}{
  andothers = {\textit{et al.},}
}

\DeclareFieldFormat[article]{volume}{\bibstring{jourvol}\addnbspace #1}
\DeclareFieldFormat[article]{number}{\bibstring{number}\addnbspace #1}
\DeclareFieldFormat[article]{volume}{Vol. #1}
\DeclareFieldFormat[article]{number}{No. #1}

\DeclareFieldFormat{url}{\bibstring{available at}\addcolon\space\url{#1}}
\DeclareFieldFormat{urldate}{\mkbibparens{accessed \addspace#1}}

\DeclareFieldFormat{urldate}{%
  \mkbibparens{accessed\space%
    \thefield{urlday}\addspace%
    \mkbibmonth{\thefield{urlmonth}}\addspace%
    \thefield{urlyear}}}

\crefformat{figure}{#2Figure~#1#3}
\Crefformat{figure}{#2Figure~#1#3}
\crefformat{table}{#2Table~#1#3}
\Crefformat{table}{#2Table~#1#3}
\crefformat{section}{#2Section~#1#3}
\Crefformat{section}{#2Section~#1#3}

\author[1]{Izabella Krzemińska}
\author[1]{Michał Butkiewicz}
\author[1]{Ewa Komkowska}

\affil[1]{Orange Research, AI Center, Warsaw, Poland
email (corr. author) Izabella.Krzeminska@orange.com}

\title{Can We Trust AI-Inferred User States? A Psychometric Framework for Validating the Reliability of User's States Classification by LLMs in Operational Environments}

\begin{document}
\maketitle

\begin{abstract}
The use of large language models to assess user states in conversational and adaptive systems is based on the assumption that the metrics used for such assessment are stable and interpretable at the level of individual scores. This paper empirically tests this assumption, focusing on the psychometric reliability of artificial intelligence (AI) measures of user states.

This study employed replication evaluation procedures to assess the repeatability of a broad set of metrics across three different bimodal large language models (GPT-4o audio, Gemini 2.0 Flash, Gemini 2.5 Flash). Analyses include both individual score reliability and aggregated reliability, allowing us to distinguish metrics potentially useful for real-time adaptation from those that retain their value only in aggregated analyses.

The results demonstrate that metric reliability cannot be considered a default property in interpretive domains. The lack of stability at the level of individual scores precludes the interpretation of such scores as indicators of user state in real-time adaptive systems, even if these metrics demonstrate stability after aggregation. At the same time, the study indicates that individually unstable metrics can retain analytical utility in post-hoc studies, identifying rules governing interactions and their relationships with user experience parameters such as satisfaction, trust, and engagement.

The main contribution of this work, besides quantifying the severity of the problem (only 31 of 213 metrics met the criteria), is the proposal of a replicable evaluation framework, enabling measurable evaluations of metric applicability. This approach supports more responsible AI design of adaptive systems, in which the interpretation of results requires explicit validation of reliability and monitoring for violations over time.

\end{abstract}

\section{Introduction}
Emotional and other human states play a central role in shaping the dynamics and outcomes of human communication, including in remote environments such as customer service call centers, where interactions are mediated almost exclusively through vocal and linguistic signals. A rich body of research has independently explored distinct modalities of emotion recognition, focusing on acoustic features (e.g., speech tempo, pitch variation) \cite{schuller2011}, textual sentiment analysis \cite{cambria2017}, and conversational behaviours such as turn-taking patterns \cite{ringeval2015}. Nevertheless, efforts to integrate these modalities into coherent predictive frameworks for customer satisfaction metrics remain at a relatively early stage of development.

The emergence of Multimodal Emotion Recognition (MMER) in Artificial Intelligence (AI) domain offers a promising avenue to bridge these gaps. Synthesizing heterogeneous emotional cues across speech, text, and interactional dynamics has demonstrated the potential to enhance recognition accuracy and system efficacy \citep{poria2017context, krzeminska2025multimodal}. Studies show that integrating acoustic, linguistic, and behavioural indicators improves the robustness of emotion-informed models for human-AI interaction across domains such as healthcare, education, and customer service \citep{wang2022systematic, ghandeharioun2019emma}. However, the application of MMER in real-time operational settings, particularly in high-demand environments like call centers, presents formidable challenges, including latency constraints, data heterogeneity, scalability issues, and critical ethical concerns such as algorithmic bias and privacy risks \citep{cowen2020, liao2021}.

Without a doubt, as the use of AI models, assistants, and agents increases, their importance in analysing user states in human-technology interactions grows. However, the lack of available research and validation of the reliability and stability of AI-generated state classifications is a challenge.

In the study by \citet{schlegel2025large}, it was shown that Large Language Models (LLMs) such as GPT‑4 and Gemini 1.5 achieve, on average, 81\% correctness in EQ (emotional intelligence) tasks, exceeding the average human score of 56\%. In particular, temporal consistency tests revealed high stability in emotional assessments between repetitions, although the models differed in their perception of subtle distinctions in emotional structure. 

In contrast,\citet{zhao2021emotion} presented an advanced MMER system based on acoustic adaptation (HuBERT) and visual feature alignment. Their model achieved a high F1 score of 0.889 in the MER2024-SEMI dataset. Despite its strong performance, the authors noted systematic differences between the results obtained under different modality configurations, arguing that consistency requires additional tuning and feature synchronization. The study did not assess the reliability of repeated input trials.

\citet{udahemuka2024multimodal}, in their review published in Applied Sciences, argues that the effectiveness of MER systems is constrained by the lack of representative datasets, insufficient contextual sensitivity, and poor robustness to missing modalities — all of which undermine the stability of emotion evaluations across modalities.

In their review, \citet{wu2025comprehensive} state that multimodal models often rely on limited, actor-generated datasets (e.g., IEMOCAP), which impedes the generalization of results to real, spontaneous interactions while artificially boosting model performance. The authors explicitly discuss the issue of cross-corpus generalizability and point to emotion classification stability across datasets as an underexplored domain in MMER research.

\citet{speer2025reliability} emphasize that while the Intraclass Correlation Coefficient (ICC) is well established in psychometric applications, it is rarely used to assess the test–retest reliability of AI systems. In their analysis of automated systems in organizational settings, ICC values often remain in the "low to moderate" range, indicating a lack of measurement repeatability in AI-based emotion classification. The authors call for systematic parallelization of trials and repetitions to confirm the reliability of AI-based tools.
Moreover, the dominant approaches to emotion recognition often rely on generalized assumptions about emotional expression, insufficiently accounting for contextual variability and individual differences \citep{barrett2017emotions}. The conceptual tension between discrete emotion models \cite{ekman2011meant} and dimensional approaches \cite{russell1991culture} further complicates the operationalisation of emotional data into actionable insights. Practical implementations frequently struggle with balancing the need for nuanced emotional detection against the operational necessity for computational efficiency and interpretability \citep{krzeminska2025multimodal}.

Therefore, a crucial question remains unresolved: \textbf{can the automatic recognition of user states inferred by LLM from real-time interaction data be reliably and accurately used for automatic experience individualization or user experience (UX) adaptation?} Preliminary findings from the existing knowledge base suggest such promise \cite{krzeminska2025multimodal}, but substantial gaps persist in terms of longitudinal stability, scalability, real-time processing capacity, and ethical robustness.

In summary, the literature indicates: (1) a lack of full test–retest reliability in AI-driven emotion classification, (2) low ICC levels in the few existing studies, (3) insufficient agreement between models and modalities despite the observed temporal consistency of emotional assessments, and (4) low predicted generalizability of current models due to dataset limitations and a lack of representativeness.

This justifies the urgent need to conduct dedicated internal reliability tests and cross-model validity assessments, which are central to the proposed methodological framework.

Therefore, tests were carried out on homogeneous material (call centre conversations) to evaluate the repeatability of user state classifications by different LLMs and to compare their outputs on the same metrics across models from various vendors.


\section{Literature Review}
The development of conversational systems based on AI has fundamentally changed the nature of human–AI interactions (HAI). A key aspect of this shift is the system's ability to construct an internal representation of the user—including their intentions, needs, communication style, and emotional state. This type of \textit{user model} forms the basis for the system’s adaptive decisions and directly influences the quality, effectiveness, and trust in HAI relationships \citep{knijnenburg2012explaining}.

Contemporary approaches to user modelling increasingly rely not on explicitly declared preferences, but on inference—deriving conclusions from contextual data, behavioural signals, or linguistic features. This opens the door to increasingly advanced hyper-personalization, but it also raises concerns about the validity, reliability, and ethics of such practices.

\citet{rai2020explainable} identifies three key components of user's trust in AI systems:
(1) Competence – the ability to perform tasks effectively;
(2) Intentionality – the belief that the system acts in the user’s best interest;
(3) Predictability – the consistency and stability of the system’s behaviour.
In this model, trust is assessed post hoc by the user.

However, in more advanced user modelling systems, accurate inference of preferences and needs can enhance the perceived competence and intentionality of the AI agent. Yet the lack of transparency in how the model makes decisions undermines predictability and may lead to trust erosion \citep{chiou2021calibrated}. This effect is amplified when personalization is based on data that the user is unaware of revealing \citep{binns2018s}.

The literature on Explainable AI (XAI) shows that transparency in user modelling affects perceptions of fairness, appropriateness, and cognitive comfort \citep{kay2006scrutable, herlocker2000explaining}. However, this effect is ambiguous. On one hand, users value understanding why the system behaves in a certain way; on the other hand, awareness of being observed can lead to behavioural change, self-presentation strategies, or even self-censorship—motivated by the desire to appear in a more favourable light (social desirability bias) \citep{joinson2001self, Eiband2018}.

\citet{Eiband2018} showed that perceived system transparency not only changes how the system is evaluated but also alters the interaction dynamics. For example, users stop responding spontaneously and begin adjusting their communication style, anticipating that their input is being analysed. In other words, they stop 'being themselves'. Can personalization to a synthetic, curated self still be beneficial?

\citet{o2008user} developed a framework of user engagement with technology, distinguishing cognitive (reflection, attention), emotional (pleasure, interest), temporal (immersion), and behavioural (intent to return) components. AI behaviours that reflect an understanding of the user can strengthen each of these engagement dimensions \citep{wen2025promoting}.

Language-based inference personalization can influence not only user experience but also cognitive style. For individuals with high Need for Cognition \cite{petty1986need}, cognitive challenges may be attractive—while for others, they may not. This raises the importance of adapting personalization strategies to both cognitive and affective profiles, as well as to contextual factors.

\citet{Lee2022} found that linguistic alignment between AI and the user increases perceived trust and satisfaction. This effect may be further enhanced by anthropomorphization, the tendency to attribute human-like traits such as intention or empathy to AI \citep{Waytz2014}. From an ethical standpoint, however, anthropomorphization can foster a false sense of AI agency, making it even more important to test the validity and reliability of systems that personalize based on inferred user states.

Traditionally, interactive systems relied on questionnaires, self-reports, or other explicit user input. Modern approaches in HAI, however, are shifting towards the automated inference of user states—emotions, motivations, communication style, cognitive engagement, and even personality traits. A prime example is the use of LLMs, which generate user state predictions based solely on text interactions without needing to ask questions like “Are you engaged?” or “Do you feel comfortable?”.

This technological shift requires a new approach to assess the quality and reliability of such mechanisms. If prediction replaces self-reports, and conversational data replaces questionnaires, we must ask ourselves:
(1) Are predictions repeatable with the same input data? (psychometric reliability); and
(2) Are predictions consistent across different LLMs? Which can be considered both as quasi-external psychometric validity (quasi since usually validated tools are used in psychometric settings \cite{zawadzki2006kwestionariusze}) or as a universality test e.g. stable repeatability across LLMs.

In a recent  experiment by \cite{cao2024large}, ChatGPT-4 was used to assess the personality traits of 226 public figures based on the TIPI (Ten Item Personality Inventory). Correlations between model-based assessments and aggregate human ratings ranged from r = 0.76 to r = 0.87—indicating that LLMs can mimic human psychometric judgment at a high level. Notably, ChatGPT-4 made these assessments in 'zero-shot mode' without prior training or feedback, demonstrating that LLMs may have embedded psychometric capacities derived from large-scale language data.

The model was reset between each item, and yet ICC across 10 repetitions ranged from 0.95 to 0.98 for each personality trait—indicating extremely high output repeatability. The alignment between ChatGPT-4 and human ratings was even stronger for more well-known individuals, measured by proxies such as Wikipedia page views.

In applied adaptive systems, user state metrics are often employed without explicit validation of their stability. Their use implicitly assumes that (i) individual metric values can be interpreted as indicators of the current user state, (ii) observed variability reflects changes in the user rather than instability in the measurement process, and (iii) such metrics can serve as reliable signals for real-time adaptation. While these assumptions are operationally convenient, they are rarely articulated or empirically examined.

In the following sections, we analyse the classes of metrics used in our study to assess reliability. Given the growing importance of personalization and transparency in HAI systems, the psychometric validation of inference mechanisms is becoming essential. This is not merely a matter of technical performance (e.g., precision or recall) but also of repeatability and interpretability from the user's perspective.

AI systems that adapt based on inferred emotional, cognitive, or communicative states must undergo the same scientific rigour as traditional measurement instruments. Only then can such systems claim epistemological credibility and legitimate authority in decision-making contexts.

Therefore, in the subsequent sections, we conduct a psychometric assessment of the repeatability of user state inferences generated by an LLM, based on experimental data under conditions of explicit and implicit personalization.

States recognition systems (including emotions) have traditionally focused on unimodal sources such as text \cite{cambria2017}, speech acoustics \cite{schuller2011}, or behavioral features like turn-taking \cite{ringeval2015}. More recent approaches integrate these signals into multimodal models leveraging psychological frameworks, including Ekman's basic emotions \cite{ekman2002facial}, Russell’s circumplex model \cite{russell1980}, and Scherer’s component process theory \cite{scherer2009dynamic}. Multimodal LLM-based models show promise for improving generalizability in real-world settings \cite{poria2017, wang2022systematic}, yet challenges remain.

Although models such as GPT-4 and Gemini 1.5 have demonstrated high average performance on emotional intelligence tasks (e.g., 81\% accuracy in \cite{schlegel2025large}), they are rarely subjected to test-retest or inter-model agreement analyses. 
Most evaluations rely on static accuracy metrics or benchmark datasets, often actor-driven and lacking real-world economical validity \cite{wu2025comprehensive}. As noted by \citeauthor{speer2025reliability}, current AI systems in organizational contexts demonstrate low-to-moderate ICCs, indicating poor repeatability of measurements \cite{speer2025reliability}.

The reviews performed by \citet{zhang2021multimodal}, \citet{udahemuka2024multimodal}, \citet{zhang2024multi}, and \citet{krzeminska2025multimodal} underscore the limitations of MMER systems that result from insufficient modality synchronization, a lack of robust datasets, and minimal testing of the stability of model output. Moreover, cross-corpus generalization remains weak due to over-fitting to narrow domains and sampling biases. These deficits are compounded by the near absence of psychometrically grounded validation strategies for emotion classification systems.

Recent work on LLM-driven user state inference increasingly explores \textit{multimodal inputs}, including both text and audio. However, explicit studies on \textit{reliability metrics} in such contexts are scarce.

\citet{speer2025reliability} examines reliability estimation for AI-generated scores and advocates for psychometric designs using ICC. Similarly, \citet{jmir2025} shows that the performance of GPT-4 in psychiatric Multiple Choice Questions (MCQ's) correlates strongly with the consistency of the test-retest. \citet{huang2024trait} demonstrates consistency in LLM personality profiling across thousands of prompts. However, most of the research lacks a true integration of voice-text and reliability evaluation.

Cyber-psychometrics extends psychometric logic to online behavior and AI-inferred states. Methods such as text or behavior mood prediction \cite{mdpi2019mood} illustrate the validity of the construct but omit stability tests. This is a limitation for systems meant to monitor user well-being over time.

Classical works such as \citet{shrout1979intraclass}, \citet{mcgraw1996inferences}, and \citet{koo2016guideline} define ICC measures applicable to repeated automated assessments. These methods are rarely used in AI-generated classification; yet, they are crucial for test–retest and measurement agreement.

\subsection{Identified Gaps and Justification}
A review of existing studies reveals four key limitations: a lack of test-retest validation for AI-based emotion classification systems, low ICC values where reliability has been tested (suggesting instability in output), inconsistent results across models and modalities despite temporal consistency, and limited generalizability due to non-representative and over-structured datasets.
The table \ref{tab:gaps} links these four gaps with implications.

\begin{table}[h!]
\centering
\begin{tabular}{@{}ll@{}}
\toprule
\textbf{Gap} & \textbf{Implication} \\ \midrule
No test--retest analysis & No evidence of measurement stability \\
Few ICC-based validations & Risk of over-interpreting raw agreement \\
Focus on construct validity only & Reliability often untested \\
Voice-text integration rarely evaluated & Real-time use cases lack empirical grounding \\
\bottomrule
\end{tabular}
\caption{Identified gaps in LLM-based user state research}
\label{tab:gaps}
\end{table}

These findings highlight a gap: most LLM-based systems do not meet the methodological standards of psychological measurement science. As noted in the foundational works \cite{shrout1979intraclass, mcgraw1996inferences, koo2016guideline}, no output variable (emotion, arousal, or behavioural intent) can be considered actionable unless its measurement process is demonstrably reliable.

We argue that AI-inferred user states must be treated as measurement constructs and validated accordingly. Our proposed framework introduces a two-level reliability assessment: (1) internal test--retest reliability via repeated application of the same model on identical inputs, and (2) inter-model agreement comparing outputs from different LLMs on the same data. Both levels employ statistical techniques such as ICC(3,1) and ICC(3,k), which are appropriate psychometrically and then correctly interpretable \cite{Brzezinski1996}.

The adoption of these measures aligns the field of affective computing with the principles of scientific measurement. It also provides a concrete operational standard for practitioners seeking to deploy adaptive AI systems in ethically sensitive and real-time environments.

\textbf{Research Hypotheses}. The main research question we ask is about the reliability of user state assessments performed by different LLMs, and three test hypotheses have been formulated.
\begin{itemize}
\item H1: User state metrics performed by LLMs are repeatable across multiple runs (measurement internal reliability).
\item H2: User state metrics performed by LLMs are consistent across different LLM (measurement inter-model validity).
\item H3: User state metrics, with an unchanged definition, remain stable when the LLM used for the measurement of inference changes.
\end{itemize}

\section{Adaptation Metric Classes for HAI Optimization}
Before discussing the test results, we summarize the class of metrics used to optimize adaptive behaviour in human-artificial intelligence (HAI) interaction.
Each class groups operational indicators that support specific adaptive goals (emotional regulation, engagement retention, relational alignment, semantic adequacy, etc.). The following taxonomy and examples consolidate the internal review and metric matrix developed as part of the project. 

\subsection{Metric classes and adaptation levers}
As formulated in the research hypotheses, the practical use of AI-based user state assessment depends not only on the availability of descriptive metrics but also on the role these metrics are expected to play within adaptive systems. In particular, the same metric may serve fundamentally different functions depending on whether it is used as a real-time adaptation signal or as an analytical indicator supporting post-hoc modelling of interaction dynamics \cite{jameson2001systems, brusilovsky2007user}.

Contemporary adaptive dialog systems, therefore, rely on heterogeneous sets of metrics operating at different temporal resolutions, levels of abstraction, and degrees of sensitivity to contextual variation. Some metrics are implicitly treated as instantaneous signals, assumed to reflect the current state of the user and directly drive adaptive responses, while others are employed exclusively in aggregated form to identify regularities in conversational structure or interaction patterns \cite{dourish2001action}.

This distinction is rarely made explicit in the literature, where metrics are more often discussed in terms of semantic plausibility or predictive utility than with respect to their functional role within adaptation pipelines and the corresponding reliability requirements \cite{hornbaek2017interaction, amershi2019guidelines}. As a result, metrics with fundamentally different constraints are frequently grouped together and used interchangeably across adaptation layers.

\paragraph{Adaptive timing.} In HAI research, the category of adaptive timing refers to how the system recognizes moments that require a change in communication strategy. Already in classic analyses of human conversations, it was observed that the course of interaction is highly sequential and that interlocutors signal transitions between dialog phases through prosodic changes, pauses, tempo, and exchange structures \cite{sacks1974simplest, goodwin1995sentence}. This concept has been developed further in works on managing dialog state, where transitions between segments of conversation are treated as a key element in regulating response strategies \citep{traum2003, young2013}.
Temporal indicators such as moments of thematic stagnation, the appearance of correction sequences, sudden shifts in user tone, shifts in turn exchange pace, or fluctuations in engagement are considered signals that should trigger a mode switch in the system. Research on conversational interaction suggests that such transitions reflect changes in the cognitive and affective states of the interlocutor and influence the further course of the conversation \cite{heldner2010pauses}. 
This work reflects contemporary temporal metrics used in HAI and (Natural Language Processing) NLP, including measures of dialog progression, stagnation detection, lack of progress (loops), drops in relevance or engagement, sudden affect shifts, and signals of topical reorganization. Therefore, adaptive timing is the way in which the system identifies the moment when a change in conversation strategy is necessary and recognizes that the current mode of dialog no longer meets the user’s needs.

\paragraph{Personalization.} The development of personalization metrics is based on the assumption that users differ in their communication and cognitive needs and preferences, which can be inferred from the course of the conversation. The very concept of the Big Five (a five-factor model of temperamental personality) is grounded in the proposed lexical assumption \cite{de201710}. Research has shown that personality traits and individual communication styles leave clear traces in language, both in vocabulary choice and in syntactic structure and content presentation \cite{krzeminska2021lexical, pennebaker2015development, mairesse2007using}. 
These studies form the basis for approaches in which adaptive systems attempt to identify long-term user preferences, such as preferred levels of detail, formality of register, or types of argumentation. 
Simultaneously, methods for profiling speech style based on statistical and semantic features have been developed and used to predict communication preferences in recommendation and dialog systems \cite{miehle2021communication}. Recently, it has also been pointed out that personalization can encompass not only personality traits but also stable thematic preferences and individual lexical patterns \cite{cao2024large}.
In the field of HAI, personalization metrics include indicators of stylistic conformity, thematic coherence, repetitive speech patterns, and signals indicating long-term user preferences. Their role is not to respond to transient states but to capture a relatively stable configuration of traits and preferences that can shape the structure of the conversation regardless of its current context.

\paragraph{Affect Alignment.} In research on affect and emotional dynamics of conversation, the foundation remains work on prosody and multimodal expression analysis. \cite{cowie2001emotion, cowie1996automatic} demonstrated that acoustic signals such as pitch, intensity, or tempo form a stable set of indicators of arousal and valence (Russell’s theory \citet{russell1980}). \cite{schuller2011}confirmed that combining these indicators with textual features leads to consistent emotional predictions in operational conditions. 
Simultaneously, \cite{cambria2017} developed affective lexicons and sentiment analysis methods based on semantic representations. These research traditions now form the basis of affective metrics used in adaptive systems both in HAI and agent systems. Prosodic and semantic signals serve as tone regulators, escalation indicators, and carriers of information about the user’s emotional dynamics.

\paragraph{Engagement.} Research on engagement, which \citet{obrien2008engagement} describe as a construct involving cognitive, emotional, and behavioral components, has long used quantitative measures expressing the quality and intensity of user participation in dialogue.
Works by \citet{ringeval2015} and \citet{ghandeharioun2019emma} show that variability in engagement can be reliably tracked using temporal parameters (response latency), utterance length, changes in turn dynamics, or the participation of reflective questions. These metrics are used in educational, therapeutic, and commercial systems as a simple yet information-rich way to monitor interaction dynamics.

\paragraph{Cognitive style.} The most complex dimension of metrics concerns cognitive style. Existing constructs in psychology, such as Need for Cognition (NFC) \cite{cacioppo1982, cacioppo1996} and Need for Cognitive Closure (NFCC) \cite{kruglanski1996}, are well-established as two independent motives for information processing. 
The first (NFC) indicates a preference for deep elaboration, while the second describes a drive to minimize uncertainty and quickly close cognitive representations. Importantly, as research by \citet{neuberg1997} and \citet{roets2011} show, NFC and NFCC do not form a single dimension and are not opposed to each other. Instead, they serve as moderators of different cognitive strategies. 
Currently, their digital equivalents are metrics of syntactic complexity, lexical density, abstraction, modality, and readability of text \cite{mcnamara2010, liao2021}, allowing generative systems to adapt the level of detail and the type of argumentation to the user’s cognitive profile.

\paragraph{Relational Synchrony.} In sociolinguistics, the central concept is the theory of communicative accommodation \cite{giles1991}, which describes the adjustment of linguistic style as a mechanism regulating interpersonal relationships. In digital data research, this phenomenon translates into relational synchronization metrics represented by syntactic and lexical similarity, linguistic politeness, register formality, or the pace of interaction coordination. \citet{danescu2011} demonstrated that style matching predicts dialogue coherence, and \citet{lee2022} confirmed the existence of similar processes in human–AI interactions.

\paragraph{Intention.} In research on user intention, key works include \citet{young2013} on Partially Observable Markov Decision Processes (POMDP) and subsequent studies by \citet{henderson2014} and \citet{mrksic2017}, which highlighted the need to measure the stability of intention classification and identify moments requiring clarification. Indicators of response consistency with intention, task phase signals, and clarification needs form the basis of task-oriented systems and regulate the model’s ability to guide the user toward achieving a goal.

\paragraph{Interactional efficiency} is a domain heavily dependent on research aimed at improving dialogues \cite{traum2003, bohus2009}. It is measured by the number of turns leading to resolution, correction frequency, semantic redundancy, and deviations from the topic. These metrics serve operational purposes, measuring and monitoring whether the system functions efficiently and predictably.

\paragraph{Semantic Appropriateness.} In the case of \textbf{semantic appropriateness}, the foundation lies in coherence and entailment metrics checking internal consistency through logical consequence or necessary implication, where the truth of one statement guarantees the truth of another. \cite{xu2018, dziri2019}, and \cite{reimers2019} describe methods for assessing whether the model’s response remains logically related to the user’s intention. In adaptive systems, these serve as warning signals, indicating potential topic drift or improper context linking.

\paragraph{Safety.} The most rapidly evolving category remains \textbf{safety} metrics. Their increasing importance reflects current times (post-LLM world) and is based on empirical observations. Since 2019, it has been shown that LLMs can generate toxic, offensive, or escalatory content with minimal prompt changes \cite{sheng2019, gehman2020, ji2023} describe hallucination risks in advisory systems. 
\cite{bender2021} highlight the repeatability of biases inherited from data, and \cite{weidinger2022}, in turn, present a systematic map of threats resulting from the model’s generativity. During the same period, legal regulations such as the EU AI Act and industry safety standards (ISO/IEC 42001:2023) emerged, making metrics of toxicity, escalation, and risk mandatory rather than optional \cite{gueorguiev2025approach}. This explains their current central role in dialogue system evaluation.

Thus, the contemporary landscape of metrics used in adaptive systems encompasses a continuum of research traditions—from transactional analysis to complex semantic models. Each of the presented classes (affect, engagement, cognitive style, relational synchronization, intention, effectiveness, semantic appropriateness, and safety) derives from rich empirical and theoretical traditions, whose understanding is necessary to assess the stability of generative models. Only on this basis can we properly define measurement scope and formulate realistic expectations for systems whose stability will be subject to further analysis.



\subsection{Theoretical background of method}
The description of the metrics used for HAI optimization provides the essential theoretical foundation for understanding what "measurement" actually entails in the context of generative systems analysing user states. However, since these measurements involve humans, it is necessary to speak of measurement in the psychometric sense. Therefore, a procedure analogous to the development of psychometric tools must be conducted, allowing for the separation of the signal (properties of the speech segment) from the fluctuations resulting from the random variability inherent in the system's operation. Psychometrics has long emphasized that a variable being measured is always a product of the properties of the object and the properties of the measurement tool \cite{cronbach1972dependability, Magnusson1971, Brzezinski1996}. 
Our assumption is that in automated systems, the properties of the computational process also play a significant role. The task of this section is thus to reconstruct, as precisely as possible, how the repeatability of measurements was evaluated, using methodology known from classical test-retest studies but adapted to new technological conditions.

In psychometric literature \cite{Brzezinski1996, zawadzki2006kwestionariusze}, a key distinction is made between variability arising from differences between analysis units and variability stemming from imperfections of the measurement instrument. In traditional human testing, the unit was the tested individual, and repeated measurements aimed to reveal the scope of random error embedded in the tool. In algorithmic analysis, the situation is different but structurally similar because the measurement unit is not a person but a segment of the recording, and repetition is not a subsequent test administration by the subject but a rerun of the same model on unchanged material. In this circumstance, the source of variability is not personal instability but "computational noise" understood as a minor generative deviations characteristic of LLMs. As noted in a section discussing reliability assessment methods, to obtain measures that can be interpreted similarly to classical psychometrics, it is necessary to apply an approach that separates these two classes of variance.

This is precisely why the methodology described in this chapter is based on a repeated measurement scheme. Each segment of the analysed material is processed multiple times by the same model under controlled identical conditions, enabling the estimation of how much of the variability in results stems from actual differences between segments and how much from the internal uncertainty of the model's operation. 
This approach aligns with the tradition of experimental psychology from the mid-20th century, where repetitions were used not to approximate mean values but to estimate the error structure. \citet{Magnusson1971} claimed that since we can never observe a "true" score directly, each measurement is simultaneously an observation and a trial to measure the accuracy of the measurement process itself.

In practice, this means that the applied procedure is not only for recording results but also for reconstructing a variance model that helps to understand the extent to which the evaluation of a given metric is stable and the extent to which it is random. 
That is why, in this study, the ICCs in the variants ICC(3,1) and ICC(3,k) were used. According to psychometric guidelines \citet{shrout1979intraclass, mcgraw1996inferences, koo2016guideline}, separating systematic variance from error variance is allowed in conditions where repetitions are treated as a fixed effect and units are considered as a source of natural differences. This approach meets the requirements of hierarchical and repeated data and simultaneously maintains, comparably to the classical test-retest scenario, even if the research paradigm has shifted from human psychology to cyber-psychometric.

It should be emphasized that the goal of this scheme is not to estimate the technical reliability of the models but to lay the groundwork for \textbf{the subsequent interpretation of user state metrics}. Only when the measurement tool demonstrates sufficient repeatability, one can analyse what the differences between segments truly mean and whether the observed metrics, independent of the class (affect, arousal, cognitive style, dialog synchrony, or semantic adequacy), meet the stability criteria necessary for their use in adaptive environments.

Taking into account the above argumentation, designed tests and data analysis were conducted based on the psychometric principle that the observed result is always a product of the properties of the measurement object and the error of the tool \citep{Magnusson1971, nunnally1975psychometric}. In the case of generative systems, this means the necessity to distinguish between two sources of variability: differences between segments of the material and fluctuations resulting from the probabilistic nature of language models. As \citet{holtzman2019curious} emphasizes, text generation by LLM-type is not a deterministic process, even with identical input. 

Even minor variations in sampling lead to noticeable semantic and pragmatic differences. This phenomenon has been confirmed in recent stability analyses of generative models, indicating that models exhibit an “inherent computational variance” independent of temperature and decoding parameters \cite{zhao2018bias}. Therefore, the analyses must assume that part of the variability in results is a feature of the model itself, rather than the studied material.

In the first stage, a variance model was built based on the assumptions of \citet{shrout1979intraclass, mcgraw1996inferences, koo2016guideline}, where a segment constitutes the unit of analysis and repetitions are treated as a fixed effect, corresponding to repeated runs of the same computational process. 
This construction allows for the separation of between-segment variance from within-segment variance, which, according to subsequent studies by \citet{mcgraw1996inferences, koo2016guideline} forms the basis for estimating reliability for repeated data. 
The choice of this particular analytical structure is also justified by the literature concerning the behaviour of language models depending on the properties of the input material. Studies by \citet{piantadosi2023modern} indicate that segment length, syntactic complexity, the number of possible interpretations, and the level of affective unambiguity influence the stability of classification in generative models.
This implies that the observed variance should be interpreted as comprising both a potentially stable component and an inferentially unstable component, in line with psychometric measurement models. Accordingly, the analysis was designed to assess the extent to which repeated evaluations under identical conditions yield a consistent signal, rather than to attribute variance to specific sources.

The second step of the procedure was a cross model analysis aimed at determining whether individual metrics are universally reliable; e.g., they are in the same class consistently across models with different architectures. Literature highlights that models trained on different datasets do not necessarily generate identical classifications, even for simple semantic categories \cite{Bommasani2021}. Therefore, comparing measurement repeatability across models serves, in this context, as an equivalent of convergent validity testing and allows for assessing whether the studied construct indicators are stable across different systems. This is important for systems using different models and provides the possibility of building solutions based on more robust metrics. 
The entire analytical procedure was shaped by the principle that measurement reliability is a necessary condition for any interpretation. As \cite{nunnally1975psychometric} reminds us, without an appropriate level of tool stability, no differences between units can be considered reliable. Regarding user state metrics, this means that only after estimating the proportion of variance can one address what the individual metric categories actually describe, whether they reflect properties of the material or are artifacts of the generative process. 

It is also worth emphasizing that although the analysis in this study concerns language models, the logic of the procedure remains close to classical psychometric methods. In traditional human tests, repeated measurement (usually performed after a short interval or by dividing the test into two equivalent parts) was used to estimate the stability of the tool, not the stability of the individual \cite{Magnusson1971, Brzezinski1996, zawadzki2006kwestionariusze}. 
\citet{nunnally1975psychometric} pointed out that internal consistency, estimated among other methods by split-half, isolates the variance component attributable to the test itself. Due to the technical possibilities in this study, repeated runs of the model on the same material serve as an equivalent to internal consistency without the need to create “halves”, but from repetitions of the generative process. The goal is not to measure the variability of the subject, but to capture the variability produced by the tool.
However, the logical structure remains analogous: if the results of two halves, i.e., two repetitions of the same process, are consistent, it indicates that the measurement instrument maintains internal coherence. In this sense, ICC(3,1) can be regarded as an equivalent of classical split-half reliability, and ICC(3,k) as its averaged version, similar to what is referred to in the literature as '“reliability of the composite score.”

Meanwhile, the cross-model analysis serves as a methodological counterpart to convergent validity testing, which assesses whether two different tools consistently measure the same construct. In psychometrics, this is often done through the correlation of results from two theoretically similar tests; in this study, the equivalent procedure involves comparing whether models with different architectures and training sets assign similar metric values to segments. Just as high correlations between methods in traditional approaches indicate that the measured construct is stable and independent of the instrument, here the consistency between models suggests that a given metric reflects a pattern present in the material rather than a random artifact.
\section{Study Design and Descriptions of Experiments}
The study consists of two distinct stages: (i) the experimental implementation stage, responsible for preparing data for analysis through controlled replication of measurements, and (ii) the analytics stage, in which stability is assessed in three different analytical scenarios.

\begin{figure}[ht]
 \centering
 \makebox[\textwidth][c]{\includegraphics[width=1\textwidth]{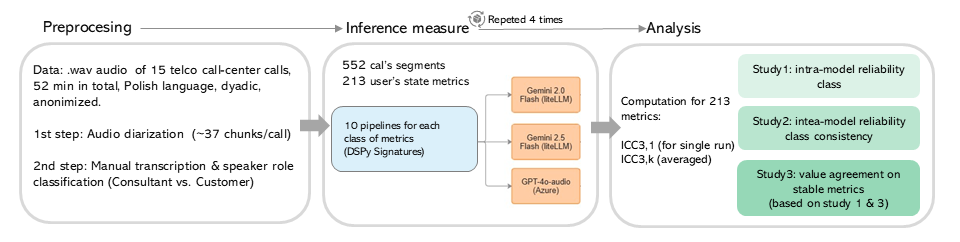}}%
 \caption{Scheme of Research Procedure}
 \label{scheme}
\end{figure}

The diagram \ref{scheme} illustrates the implementation of the experiment at the operational level, i.e., the process of preparing the research material and generating input data for analysis. Its purpose is to present the necessary data regarding the amount of material studied, to enable the replication of the procedure, and to control the conditions under which the data used in all three studies were generated.

\subsection{The Experimental Stage}
\paragraph{Dataset Composition}
The stability test was conducted by running ten pipelines with commands to measure respective set of metrics through an identical dataset four times. For the immediate utility of the metric tests, the material used for the tests was the target material, i.e., telephone calls from a contact centre.

The test material included audio recordings and transcripts of 15 anonymized dyadic calls in the call centre with a total duration of 52 minutes (average 3.5 minutes per call), resulting in 552 recorded segments for analysis and an average of approximately 37 segments per call seee Fig. \ref{scheme}.


Preparing accurate transcriptions and not relying on automated tools eliminated any bias resulting from differences in transcription quality between the models. The goal was to create the best possible isolated, identical conditions for measuring user state metrics for three different LLM models.

Due to the measurement objective and the desire to evaluate the models' performance both in real time (segment) and globally for post-analysis of the entire conversation, and to verify metrics that could be used in the RTw call center, three models were selected that supported audio and text in Polish: Gemini 2.0 Flash, Gemini 2.5 Flash \cite{gemini}, and GPT-4o audio \cite{gpt4o}.


\paragraph{Pipeline details}
All three models were evaluated under identical experimental conditions to ensure comparability. Inference was conducted using the DSPy framework\footnote{https://dspy.ai/} (version 2.6.23), which programmatically generates prompts from declarative signatures rather than using raw prompt strings \cite{dspy1, dspy2}. Each model received inputs structured through DSPy modules that specified the required metrics and expected output format; these inputs comprised audio recordings (.wav) and their corresponding text transcriptions. We provide comprehensive descriptions of each metric and expected output format in Table \ref{tab:metrics_full} in the Appendix.
Inference parameters were standardised where appropriate to ensure a fair comparison of stability. The temperature was set to 0.3 across all models to minimise variance while maintaining coherent generation. Secondary sampling parameters (such as top-p and top-k) were maintained at their respective manufacturer-recommended default values (top-p=0.95 for Gemini, top-p=1.0 for GPT-4o) to respect the distinct architectural optimisations of each model and adhere to provider guidelines against simultaneous parameter tuning. The maximum token limit was set to 8,192 (constrained by Gemini 2.0 Flash limits). DSPy's caching functionality was disabled to ensure independent inference runs. Each sample was processed four times to assess output stability using the following model versions: GPT-4o Audio (gpt-4o-audio-preview-2024-12-17), Gemini 2.0 Flash (gemini-2.0-flash-001), and Gemini 2.5 Flash (gemini-2.5-flash). The input specification included descriptions of all metrics to be measured and their expected value constraints. Post-processing validation was applied to all model outputs. Responses violating these constraints were excluded from the results.

\subsection{The Analytics Stage}
The analytical stage consists of three complementary studies, each examining a different aspect of metric stability relevant to practical applications in adaptive systems. All analyses were conducted on a single dataset obtained using the procedure described above (\ref{scheme}).

The common goal of all three studies is to assess whether user state metrics can be considered reliable indicators under various usage conditions. Stability is not understood as an abstract property of a model, but as an empirically observable characteristic of metric results when identical input conditions are repeatedly assessed.

The first study examines the stability of user state metrics within a single model, with controlled repeatability. The goal is to determine whether repeated assessments of identical input conditions produce sufficiently consistent metric values to justify their interpretation at the individual output level. In a single sentence, we seek to answer the question: Does a given LLM model measure a given metric stably? We apply psychometric criteria for metric goodness of fit.

Study 1 addresses the most basic assumption: that a single metric value can be meaningfully interpreted as an indicator of the user's current state. Stability is assessed by comparing repeatable metric results and assessing the proportion of metrics meeting predefined reliability thresholds.

The results provide a baseline characterization of metric stability and reveal the extent to which stability can be considered a default feature or an exception among commonly used metrics (referring to H1).

The second study focuses on verifying the universality of the assigned interpretation classes between models. We verify whether the class measured in Study 1 is invariant for the metric and independent of the model used for evaluation. In other words, we examine the metric's robustness to model changes, including potentially changes not reflected in subsequent model releases. Hence, we compare models from different vendors and two available models from a single vendor, but from subsequent generations. Study 2 answers the question of whether the metric class assigned in Study 1 is universal and robust to changes in the LLM model (referring to H2 and H3).

The third study addresses the question of whether the metrics we considered universally stable in Studies 1 and 2 generate nearly identical results for the same input data (referring to H2 and H3).
Thanks to this multi-layered analytical design, it is possible not only to determine whether the metrics are stable (Study 1) but also to observe the degradation of their universality with the addition of further qualitative conditions (repeatability, universality (study 2: cross-model universality of the reliability and study 3: cross-model consistency of result). The second, ICC(3,k), allows for verifying the metric's suitability for global analyses, such as post-facto final evaluations, analyses aimed at building a hierarchy of adaptive systems, or simply for user insights.

\paragraph{Metrics of Reliability and Their Interpretation} 
Reliability is analysed separately using ICC(3,1), capturing the stability of a single inference, and ICC(3,k), representing the upper bound of stability achievable through aggregation across repeated runs. This separation allows the instability intrinsic to the metrics to be distinguished from the variance reduction effects introduced by averaging and replication \cite{shrout1979intraclass}.
Both measures are relevant in different application contexts: ICC(3,1) is critical for real-time scenarios such as adaptive user experience systems, whereas ICC(3,k) is appropriate for offline analyses of conversational patterns and for evaluating the contribution of metrics to global outcome measures, such as overall contact assessment, engagement, overall trust, or Net Promoter Score (NPS).

In the tests we used the following thresholds for each interpretation class using methodological background from \cite{shrout1979intraclass}: \\
\textit{Excellent reliability} - ICC value is between 0.9 and 1.0 \\
\textit{Good reliability} - ICC value is greater or equal to 0.75 and smaller than 0.9 \\
\textit{Moderate reliability} - ICC value is greater or equal to 0.5 and smaller than 0.75 \\
\textit{Poor reliability} - ICC value is smaller than 0.5

Additionally, the analysis of inter-model agreement in metric values is meaningful only for metrics with empirically confirmed stability, as otherwise the observed differences may be entirely dominated by random variance of single inferences rather than by systematic measurement differences. For the same reason, investigating the effects of user-state measurement performed by LLMs is not justified without prior verification of reliability, which constitutes a necessary precondition for any comparative interpretation. Restricting the analysis to stable metrics therefore makes it possible to disentangle measurement instability from genuine agreement or disagreement of metric values across models.

\paragraph{Cross-model comparison}
The use of the three evaluated models enables two complementary comparison regimes grounded in distinct methodological assumptions. 
The first regime is a vertical, within-family comparison between successive versions of the same model, namely Gemini 2.0 Flash and Gemini 2.5 Flash. This comparison supports the analysis of changes in metric reliability within a shared architectural and training lineage, under the assumption of a consistent modeling philosophy but differing scale, training data coverage, and alignment procedures. Within this regime, observed differences in ICC values are attributed to model evolution rather than to differences between model classes.

The second regime is a horizontal, cross-family comparison between models belonging to the same operational class but originating from different architectural families, namely Gemini 2.5 Flash and GPT-4o audio. The purpose of this comparison is not model ranking but the identification of differences in variance structures and reliability profiles of metrics arising from distinct response generation mechanisms, training strategies, and modality handling.

In addition, intra-model comparisons across the three multimodal models are used to assess the degree of metric universality and their robustness to model evolution and architectural variation.

\section{Proposed Evaluation Metrics and their interpretation}
\subsection{Reliability metrics for repeated measures}
In assessing the stability of AI-inferred user state metrics, several approaches from psychometrics were considered. 
To assess the internal reliability of user state metrics inferred by LLMs, we implemented a repeated measures design in which the same segments of conversation were processed multiple times by the same model. This procedure evaluates the consistency of the results  under identical input conditions and provides a psychometrically valid estimate of measurement stability. 

The unit of analysis was not the user, but the \textit{segment} of a recorded utterance. Each segment was processed repeatedly (four times: A--D) by the same script under identical conditions. Reliability was operationalized as the repeatability of the same variable (e.g. arousal, valence) for a given segment across those repetitions. This corresponds to the classical test--retest scenario, adapted to technical systems instead of human traits (cf. \cite{shrout1979intraclass, koo2016guideline}).

The most appropriate indicators are the \textbf{Intraclass Correlation Coefficients ICC(3,1) and ICC(3,k)}, which model the measurement variance in repeated assessments under fixed conditions \citep{shrout1979intraclass, mcgraw1996inferences, koo2016guideline}. 
ICC(3,1) reflects the repeatability of a single model output, while ICC(3,k) evaluates the stability of the mean in k replications. Both coefficients are interpretable on the scale $0$–$1$, with thresholds $\geq 0.75$ indicating good and $\geq 0.90$ very good reliability \citep{koo2016guideline}. These indices are consistent with the \textit{test--retest paradigm} and allow separating meaningful between-segment variance from within-segment error.

\textbf{ICC(3,1)} evaluates the reliability of a single run by quantifying the proportion of variance attributable to systematic differences between segments rather than random error:
\[
ICC(3,1) = \frac{MS_B - MS_W}{MS_B + (k-1)MS_W}
\]
where $MS_B$ is the mean square between targets (segments), $MS_W$ the within-target variance across replications, and $k$ the number of replications. 

\textbf{ICC(3,k)} extends this logic to the mean of $k$ replications, providing a more robust estimate of the measurement stability:
\[
ICC(3,k) = \frac{MS_B - MS_W}{MS_B}
\]
Alternative measures were considered but ultimately rejected. \textbf{Pairwise correlations} (e.g. Pearson's $r$, Spearman's $\rho$) only capture covariation and may inflate reliability estimates despite systematic shifts between replications; moreover, they do not measure absolute agreement and become cumbersome with more than two repetitions. \textbf{Internal consistency indices} such as \textbf{Cronbach's $\alpha$} or \textbf{McDonald's $\omega$} assume the independence of items and a latent construct underlying multiple test items \citep{Brzezinski1996,Magnusson1971}. These assumptions are violated in the present context, where repetitions (A--D) represent technical replications of the same segment, not independent items. Applying $\alpha$ or $\omega$ would therefore misrepresent measurement error and overstate reliability.
For these reasons, \textbf{ICC(3,1) and ICC(3,k) were selected as the primary reliability indicators}, as they best reflect the hierarchical and repeated-measurement nature of AI-based inference on call center data and provide psychometrically valid, interpretable evidence of measurement stability.

\subsection{Practical interpretation of ICC(3,1) vs ICC(3,k).}
In adaptive HAI systems, the choice between ICC(3,1) and ICC(3,k) has practical implications. 
\textbf{ICC(3,1)} reflects the repeatability of a single model output under fixed conditions. It answers the question: \textit{``How consistent is a single run of the model on the same input segment?''} This is the stricter criterion and directly reflects the stability of raw predictions available in operational, single-pass settings.

By contrast, \textbf{ICC(3,k)} reflects the reliability of an averaged score across $k$ repeated runs. Averaging reduces random error (the within-segment variance $MS_W$ is divided by $k$), which mechanically inflates the coefficient. As a result, ICC(3,k) values are typically \textbf{substantially higher} than ICC(3,1), often approaching the “very good” range even when single-run reliability remains only moderate. This is not a methodological artifact but a direct mathematical property: aggregation smooths out stochastic fluctuations of the model.

For this reason, \textbf{results should be interpreted separately for ICC(3,1) and ICC(3,k)}. High ICC(3,k) indicates that stability can be achieved through repeated inference or ensemble averaging, which may be feasible in offline analytics but less so in real-time systems. In turn, ICC(3,1) provides a more realistic assessment of reliability in single-shot operational scenarios, such as live call centre support. 

In practice, reporting both values provides a complete picture: (1) the baseline repeatability of raw model outputs (ICC(3,1)) and (2) the upper bound of achievable stability under aggregation strategies (ICC(3,k)).

\subsection{Consistency metrics for cross model evaluation }

The choice of agreement measures depended on the type of output scale of the given feature (continuous, ordinal, binary, or nominal). As the subset that demonstrated excellent single-inference reliability contained only continuous and binary features, the following agreement statistics were used: for the continuous scale, the Mean Absolute Error (MAE) was selected, while for the binary scale, Cohen's Kappa was used.

\subsection{Inter-model consistency of metric values: binary and continuous metrics}

To assess inter-model consistency of metric values (Study~3), distinct analytical procedures were applied to binary/categorical and continuous metrics, reflecting their different measurement properties. The analysis was restricted to metrics exhibiting excellent internal stability (ICC $\geq 0.9$), ensuring that observed discrepancies reflect inter-model differences rather than stochastic variability of unstable measurements.

For binary and categorical metrics, inter-model consistency was evaluated using \textbf{Cohen’s kappa coefficient} ($\kappa$), which quantifies the agreement between two raters beyond chance level.

Let $y^{A}_{s}$ and $y^{B}_{s}$ denote the categorical outputs of two models $A$ and $B$ for segment $s \in \{1,\dots,S\}$. Cohen’s kappa is defined as:

\begin{equation}
\kappa = \frac{p_o - p_e}{1 - p_e},
\end{equation}

where $p_o$ is the observed proportion of agreement between models and $p_e$ is the expected agreement by chance.

Kappa values were interpreted according to established guidelines proposed by Landis and Koch (1977) and Fleiss (1981), which classify agreement strength into discrete categories ranging from poor to near-perfect agreement. These thresholds provide a well-established reference framework for evaluating inter-model consistency of categorical interpretations.

For continuous metrics, inter-model differences were quantified using the \textbf{mean absolute error} (MAE), which measures the average magnitude of discrepancy between model outputs across conversation segments.

Let $x^{A_i}_{s}$ and $x^{B_j}_{s}$ denote the values of a continuous metric for segment $s \in \{1,\dots,S\}$ obtained from run $i$ of model $A$ and run $j$ of model $B$, respectively. For each pair of runs $(i,j)$, MAE was computed as:

\begin{equation}
\mathrm{MAE}_{i,j} = \frac{1}{S} \sum_{s=1}^{S} \left| x^{A_i}_{s} - x^{B_j}_{s} \right|.
\end{equation}

To account for generative variability, all combinations of repeated runs were considered ($i,j \in \{1,\dots,4\}$), yielding 16 MAE values per model pair. The typical inter-model discrepancy was estimated as the median across run-pair combinations:

\begin{equation}
\widetilde{\mathrm{MAE}} = \mathrm{median}_{i,j} \left( \mathrm{MAE}_{i,j} \right).
\end{equation}

Because continuous metrics differed in scale and numerical range, MAE values were normalized by the empirical range of each metric:

\begin{equation}
\mathrm{nMAE}_{\text{range}} =
\frac{\widetilde{\mathrm{MAE}}}{\max(x) - \min(x)},
\end{equation}

where $\max(x)$ and $\min(x)$ denote the maximum and minimum observed values of the metric across all segments. The resulting $\mathrm{nMAE}_{\text{range}}$ is a dimensionless quantity bounded in $[0,1]$, enabling comparison across metrics with heterogeneous scales.

For the purpose of unified reporting and synthesis of results across binary and continuous metrics, discrete classes of inter-model comparability were defined.

For categorical metrics, comparability classes were derived from $\kappa$ values following established interpretative scales. For continuous metrics, operational thresholds were defined \emph{a priori} for $\mathrm{nMAE}_{\text{range}}$, reflecting practical levels of inter-model agreement:
\begin{itemize}
    \item $\mathrm{nMAE}_{\text{range}} \leq 0.05$ — \textbf{near-ideal agreement};
    \item $0.05 < \mathrm{nMAE}_{\text{range}} \leq 0.10$ — \textbf{moderate agreement};
    \item $0.10 < \mathrm{nMAE}_{\text{range}} \leq 0.20$ — \textbf{low agreement};
    \item $\mathrm{nMAE}_{\text{range}} > 0.20$ — \textbf{non-acceptable agreement / lack of agreement}.
\end{itemize}

These thresholds should be interpreted as operational decision boundaries for assessing inter-model comparability of metric values, rather than as universal psychometric norms. Together, $\kappa$ and normalized MAE provide complementary perspectives on inter-model consistency, capturing agreement in categorical interpretations and magnitude of discrepancy in continuous estimates, respectively.

\section{Reliability (Test–Retest) Results}
In our tests, ICC(3,1) values (single-run reliability) already reached “good” to “excellent” levels (0.93–0.99 for several metrics). However, ICC(3,k) values (reliability of the mean across four replications) approached unity (0.999+). This discrepancy is expected: averaging reduces within-segment error variance, thereby inflating reliability estimates. While ICC(3,k) demonstrates the upper bound of stability achievable under aggregation, ICC(3,1) provides a realistic picture of reliability in single-pass operational settings.

For this reason, metrics should be compared and interpreted separately for ICC(3,1) and ICC(3,k). High ICC(3,k) indicates that stable estimates can be achieved by ensemble averaging or repeated inference, which may be feasible in offline analytics but less applicable in real-time systems.

At the same time, the measurement results indicate a non-uniform character of this improvement. For example, within the Personalization metric class, a non-obvious pattern is observed: these metrics exhibit high and stable ICC(3,1) reliability in the Gemini 2.0 Flash model, whereas in Gemini 2.5 Flash their average stability decreases substantially, despite comparable ICC(3,k) values. This indicates that the decline in reliability primarily affects single inferences rather than the potential for stabilization through aggregation. While this pattern requires further in-depth analysis, it provides evidence that, as a model evolves, the sensitivity of certain metric classes to generative fluctuations may change, even within the same architectural family.

\begin{table}
\small
\centering
\caption{Gemini 2.0 Flash reliability by metric class for ICC(3,1): min, max, mean ICC and counts of `excellence' (\(\geq 0.90\)) and `good' (\([0.75, 0.90]\)) metrics for ICC(3,1).
Total number of calculated metric N=203 (14 were not calculated)}
\label{tab:Gem 2.0 icc_1_class_summary}
\begin{tabular}{lrrrrr}
\toprule
metric class (N) & min & max & mean & excellence n & good n \\
\midrule
adaptive (11) & 0.38 & 0.85 & 0.57 & 0 & 3 \\
affect\_alignment(60)  & -0.02 & 0.99 & 0.50 & 3 & 16 \\
cognitive\_style (23)  & 0.10 & 0.74 & 0.50 & 0 & 0 \\
engagement (46)  & 0,00 & 1 & 0.67 & 18 & 9 \\
intention (8) & 0.55 & 1 & 0.90 & 7 & 0 \\
interactional\_efficiency (14)  & 0.33 & 0.73 & 0.51 & 0 & 0 \\
personalization (13)  & 0.90 & 0.99 & 0.98 & 12 & 1 \\
relational\_synchrony (11) & 0.75 & 0.99 & 0.89 & 6 & 5 \\
safety (2) & 0.82 & 0.87 & 0.85 & 0 & 2 \\
semantic\_appropriateness (15) & 0.30 & 0.92 & 0.79 & 3 & 8 \\
\bottomrule

\end{tabular}
\end{table}

\begin{table}
\centering
\small
\caption{Gemini 2.5 Flash reliability by metric class for ICC3,1: min, max, mean ICC and counts of `excellence' (\(\geq 0.90\)) and `good' (\([0.75, 0.90]\)) metrics for ICC(3,1). 
Total number of calculated metric N=197 (17 were not calculated)}
\label{tab:Gem 2.5 icc_1_class_summary}
\begin{tabular}{lrrrrr}
\toprule
metric class (N) & min & max & mean & excellence n & good n \\
\midrule
adaptive (11) & 0.97 & 1 & 0.99 & 11 & 0 \\
affect\_alignment(63)  & -0.05 & 0.99 & 0.45 & 4 & 14 \\
cognitive\_style (14)  & 0.999 & 1 & 1 & 7 & 7 \\
engagement (45)  & 0,77 & 1 & 0.99 & 45 & 1 \\
intention (8) & 0.99 & 1 & 1 & 8 & 0 \\
interactional\_efficiency (14)  & 0.999 & 1 & 1 & 14 & 0 \\
personalization (12)  & 0.56 & 0.99 & 0.85 & 4 & 5 \\
relational\_synchrony (12) & 0.88 & 0.99 & 0.93 & 11 & 1 \\
safety (3) & 0.86 & 0.98 & 0.91 & 2 & 1 \\
semantic\_appropriateness (15) & 1 & 1 & 1 & 15 & 0 \\
\bottomrule
\end{tabular}
\end{table}

\begin{table}
\small
\centering
\caption{GPT-4o audio reliability by metric class for ICC3,1: min, max, mean ICC and counts of `excellence' (\(\geq 0.90\)) and `good' (\([0.75, 0.90]\)) metrics for ICC(3,1). 
Total number of calculated metric N=200 (15 were not calculated)}
\label{tab:GPT 4o icc_1_class_summary}
\begin{tabular}{lrrrrr}
\toprule
metric class (N) & min & max & mean & excellence n & good n \\
\midrule
adaptive (11) & 0.37 & 0.99 & 0.709 & 4 & 2 \\
affect\_alignment(63)  & 0.17 & 1 & 0.78 & 24 & 19 \\
cognitive\_style (15)  & 0 & 0,99 & 0.75 & 7 & 3 \\
engagement (43)  & 0.12 & 1 & 0.84 & 22 & 12 \\
intention (7) & 0.31 & 1 & 0.78 & 3 & 2 \\
interactional\_efficiency (15)  & 0.65 & 0.99 & 0.88 & 9 & 4 \\
personalization (12)  & 0.78 & 0.96 & 0.88 & 6 & 6 \\
relational\_synchrony (17) & 0.77 & 0.96 & 0.85 & 6 & 11 \\
safety (3) & 0.65 & 0.84 & 0.76 & 0 & 2 \\
semantic\_appropriateness (14) & 0.54 & 1 & 0.85 & 8 & 3 \\
\bottomrule
\end{tabular}
\end{table}

\subsection{Study 1 results: intra model test-retest reliability}
The first study examines model reliability, understood as the consistency of repeated queries issued on the same material and quantified using the ICC(3,1) and ICC(3,k). Features with insufficient variance (dominant category prevalence $\geq99\% $) were excluded prior to analysis to ensure robust stability estimation. The specific exclusion counts for each model are reported in the respective table descriptions.

Tables Tab \ref{tab:Gem 2.0 icc_1_class_summary}, Tab \ref{tab:Gem 2.5 icc_1_class_summary}, Tab \ref{tab:GPT 4o icc_1_class_summary}, Tab \ref{tab:Gemi 2.0 icc_k_class_summary}, Tab \ref{tab:Gem 2.5 icc_k_class_summary}
Tab \ref{tab:GPT 4o icc_k_class_summary} reveal a systematic difference between stability estimated for single inferences and stability achieved through aggregation across repeated runs. For all analysed models, a substantial increase in reliability is observed when moving from ICC(3,1) to ICC(3,k), confirming the reduction of within-segment variance through averaging.

At the same time, the distributions of ICC(3,1) values exhibit pronounced heterogeneity across metric classes, indicating that metric stability is strongly dependent on the nature of the metric itself or on the category of the conversational state being measured.

The vertical comparative analysis of successive versions of the Gemini model reveals an overall increase in the number of metrics reaching good and excellent reliability levels for ICC(3,1) in version 2.5 compared to version 2.0. This improvement is particularly evident in metric classes such as Engagement, Interactional Efficiency, and Semantic Appropriateness, where the reliability distributions shift markedly toward higher values.
\begin{table}
\small
\centering
\caption{Gemini 2.0 Flash reliability by metric class for ICC(3,k): min, max, mean ICC(3,k) and counts of `excellence' (\(\geq 0.90\)) and `good' (\([0.75, 0.90]\)) metrics for ICC(3,k). 
Total number of calculated metric N=200 (14 were not calculated)}
\label{tab:Gemi 2.0 icc_k_class_summary}
\begin{tabular}{lrrrrr}
\toprule
metric class (N) & min & max & mean & excellence n & good n \\
\midrule
adaptive (11) & 0.71 & 0.96 & 0.83 & 3 & 6 \\
affect\_alignment(60)  & -0.08 & 0.99 & 0.70 & 22 & 18 \\
cognitive\_style (23)  & 0.30 & 0.92 & 0.77 & 3 & 17 \\
engagement (46)  & 0 & 0.99 & 0.79 & 27 & 8 \\
intention (8) & 0.83 & 1 & 0.97 & 7 & 1 \\
interactional\_efficiency (14)  & 0.67 & 0.91 & 0.80 & 1 & 8 \\
personalization (13)  & 0.97 & 1 & 0.99 & 13 & 0 \\
relational\_synchrony (11) & 0.92 & 0.99 & 0.97 & 11 & 0 \\
safety (2) & 0.95 & 0.97 & 0.96 & 2 & 0 \\
semantic\_appropriateness (15) & 0.64 & 0.98 & 0.93 & 11 & 3 \\
\bottomrule
\end{tabular}
\end{table}

\begin{table}
\small
\centering
\caption{Gemini 2.5 Flash reliability by metric class for ICC(3,k): min, max, mean ICC(3,k) and counts of `excellence' (\(\geq 0.90\)) and `good' (\([0.75, 0.90]\)) metrics for ICC(3,k). 
Total number of calculated metric N=197 (17 were not calculated)}
\label{tab:Gem 2.5 icc_k_class_summary}
\begin{tabular}{lrrrrr}
\toprule
metric class (N) & min & max & mean & excellence n & good n \\
\midrule
adaptive (11) & 0.992 & 0.999 & 0.996 & 11 & 0 \\
affect\_alignment(63)  & -0.25 & 0.996 & 0.63 & 23 & 8 \\
cognitive\_style (14)  & 0.999 & 1 & 0.999 & 14 & 0 \\
engagement (45)  & 0.93 & 1 & 0.99 & 45 & 0 \\
intention (8) & 0.99 & 1 & 0.999 & 8 & 0 \\
interactional\_efficiency (14)  & 0.999 & 1 & 0.999 & 14 & 0 \\
personalization (12)  & 0.83 & 0.99 & 0.95 & 11 & 1 \\
relational\_synchrony (12) & 0.97 & 0.99 & 0.98 & 12 & 0 \\
safety (3) & 0.96 & 0.99 & 0.98 & 3 & 0 \\
semantic\_appropriateness (15) & 1 & 1 & 1 & 15 & 0 \\
\bottomrule
\end{tabular}
\end{table}

\begin{table}
\small
\centering
\caption{GPT-4o audio reliability by metric class for ICC(3,k): min, max, mean ICC(3,k) and counts of `excellence' (\(\geq 0.90\)) and `good' (\([0.75, 0.90]\)) metrics for ICC(3,k). 
Total number of calculated metric N=200 (15 were not calculated)}
\label{tab:GPT 4o icc_k_class_summary}
\begin{tabular}{lrrrrr}
\toprule
metric class (N) & min & max & mean & excellence n & good n \\
\midrule
adaptive (11) & 0.93 & 0.99 & 0.96 & 11 & 0 \\
affect\_alignment(58)  & 0.34 & 0.99 & 0.84 & 28 & 17 \\
cognitive\_style (23)  & 0.93 & 0.99 & 0.97 & 23 & 0 \\
engagement (46)  & 0.80 & 1 & 0.96 & 42 & 4 \\
intention (8) & 0.89 & 0.99 & 0.97 & 7 & 1 \\
interactional\_efficiency (13)  & 0 & 1 & 0.89 & 12 & 0 \\
personalization (12)  & 0.999 & 0.999 & 0.999 & 12 & 0 \\
relational\_synchrony (12) & 0.96 & 0.99 & 0.98 & 12 & 0 \\
safety (2) & 0.95 & 0.96 & 0.96 & 2 & 0 \\
semantic\_appropriateness (15) & 0 & 1 & 0.89 & 12 & 2 \\
\bottomrule
\end{tabular}
\end{table}

In contrast, the horizontal comparison between models of the same operational class but based on different architectures—namely Gemini 2.5 Flash and GPT-4o audio—reveals differences not so much in the maximum attainable ICC(3,k) values, which are high for most metric classes in both models, but rather in the structure of ICC(3,1) variance across metric classes. The GPT-4o audio model is characterized by a more uniform distribution of ICC(3,1) reliability across metric classes, resulting in smaller differences between areas of high and low stability. By contrast, Gemini 2.5 Flash exhibits greater heterogeneity: alongside metric classes with very high stability (e.g., Relational Synchrony, Personalization), there are areas with clearly reduced single-metric reliability, particularly in Affect Alignment and Adaptive.

The measurements indicate a recurring pattern whereby differences between Gemini 2.5 Flash and GPT-4o audio are most pronounced for metrics of an interpretative and semantically complex nature, whereas deterministic metrics (e.g., count-based and length-based indicators) exhibit high and comparable stability in both models. Generalizing this pattern as a principle would require testing across a broader range of models.

Based on the obtained measurements, it can be concluded that:
(1) a higher model version does not lead to a uniform improvement in stability across all metric classes;
(2) differences between models of the same operational class are manifested primarily in the variance structure of ICC(3,1), rather than in the upper-bound values of ICC(3,k); and
(3) the character of the metric (deterministic vs. interpretative) may constitute a key factor differentiating inference stability, independently of the model.

\subsection{Study 2 results: Universality of Metric Repeatability Across Models}

The second study focuses on assessing the universality of metric repeatability across models, understood as the preservation of the same psychometric reliability class for a given metric when the model is changed. To this end, inter-model reliability class correspondence plots are employed, enabling pairwise comparisons of models with respect to metric stability classes.

\begin{figure}[ht!]
 \centering
 \makebox[\textwidth][c]{\includegraphics[width=1\textwidth]{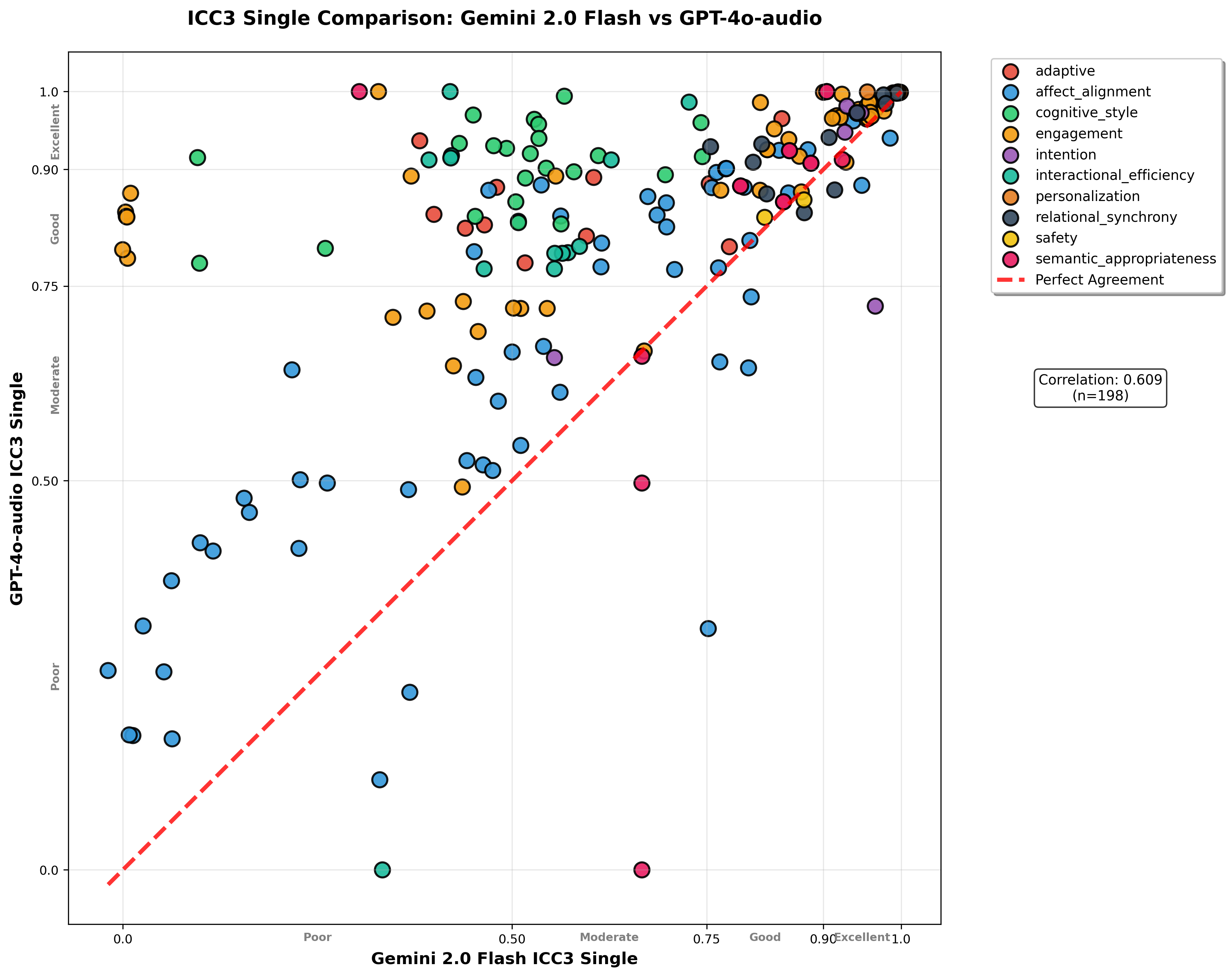}}%
 \caption{Matrix of metrics - ICC(3,1) single comparison between GPT-4o audio and Gemini 2.0 Flash}
 \label{fig-1}
\end{figure}

\begin{figure}[ht]
 \centering
 \makebox[\textwidth][c]{\includegraphics[width=1\textwidth]{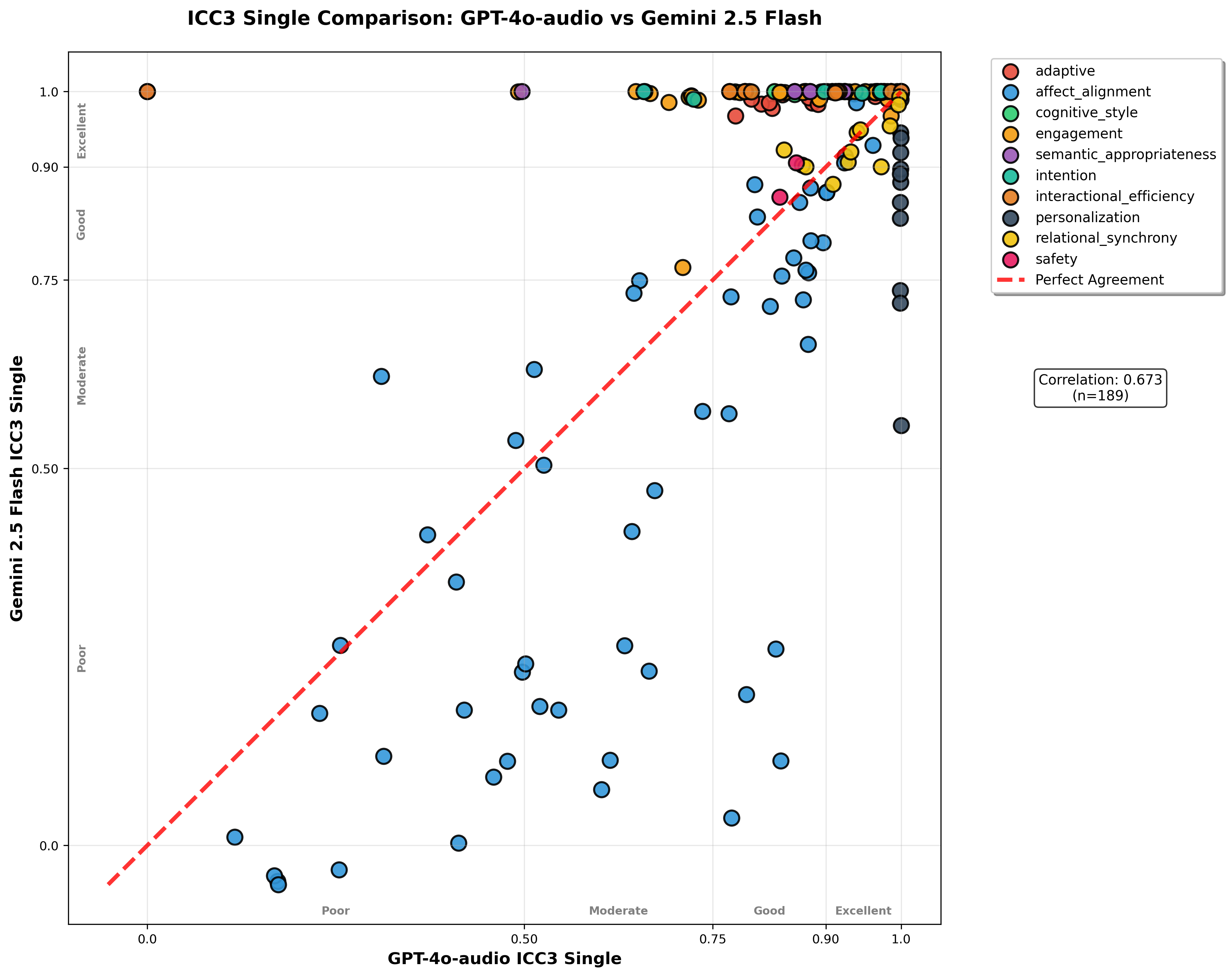}}%
 \caption{Matrix of metrics - ICC(3,1) single comparison (between GPT-4o audio and Gemini 2.5 Flash}
 \label{fig-2}
\end{figure}

\begin{figure}[ht!]
 \centering
 \makebox[\textwidth][c]{\includegraphics[width=1\textwidth]{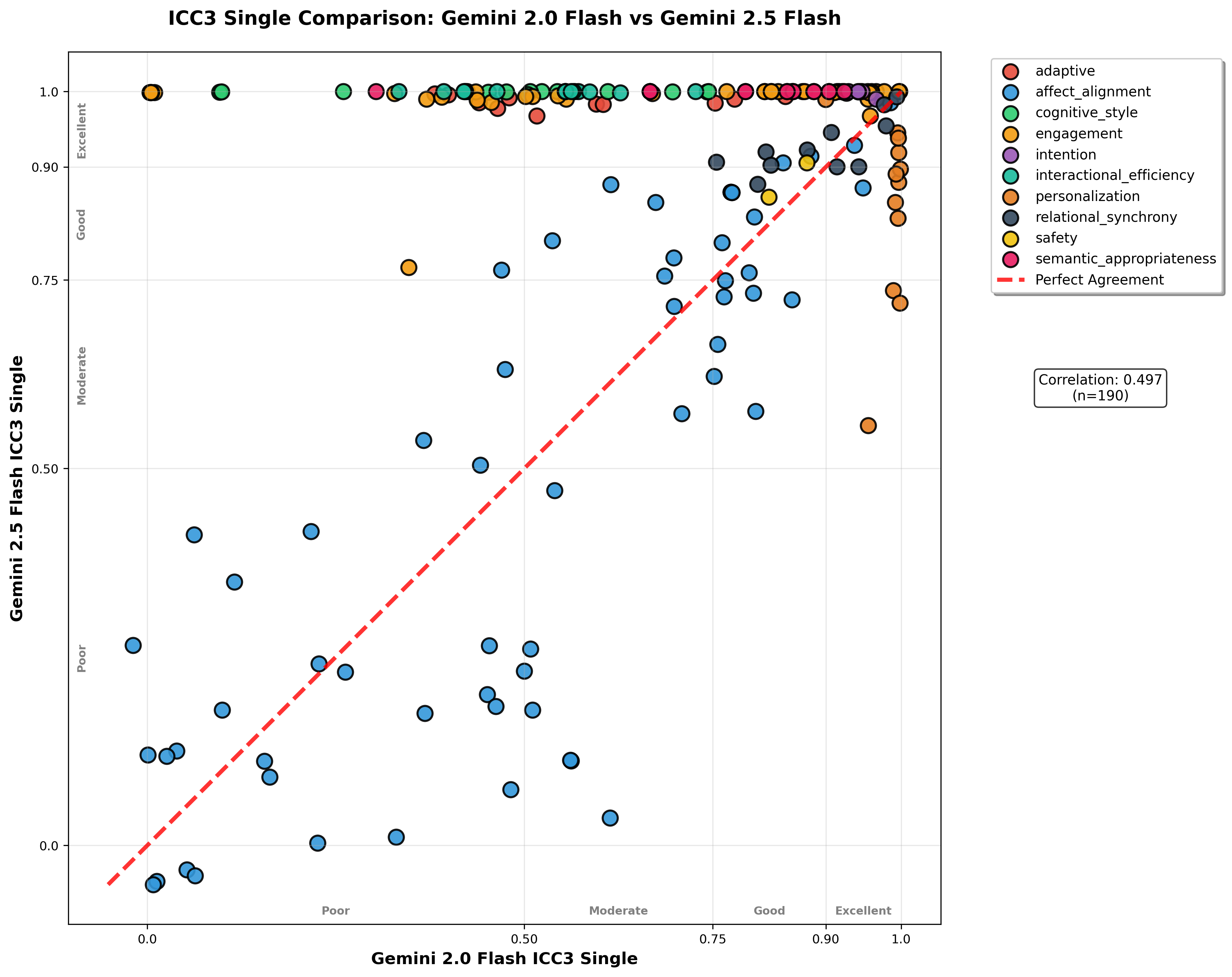}}%
 \caption{Matrix of metrics - ICC(3,1) single comparison between Gemini 2.0 Flash and Gemini 2.5 Flash}
 \label{fig-3}
\end{figure}

In each of the three figures (Fig,\ref{fig-1}, Fig.\ref{fig-2}, Fig. \ref{fig-3}) the axes represent psychometric reliability classes of metrics, determined on the basis of ICC coefficient values, for the two models being compared. Each point in the plot corresponds to a single metric, with point colour indicating the metric’s membership in a specific pipeline. Points located along the diagonal indicate metrics whose reliability class is preserved regardless of model choice, whereas off-diagonal points reflect shifts in the stability class of the same metric between models.

The proposed analysis makes it possible to identify metrics that are robust to architectural changes and model evolution, as well as metrics that are sensitive to the choice of a particular model or its version. This has a twofold practical significance. First, it enables informed model selection in relation to metrics that are critical for a given use case. Second, it supports the selection of metric sets for deployment in multi-model systems, where measurement comparability and consistency are of central importance.

In addition, the analysis of reliability class correspondence provides guidance for the design of inference aggregation and replication strategies by indicating which metrics require repeated executions in order to achieve an acceptable level of stability under conditions of model heterogeneity.

Analysis of inter-model reliability class correspondence plots reveals that only a small number of metrics remain in the highest psychometric reliability class $\mathrm{ICC}(3,1) \ge 0.9$ for both compared models simultaneously. Most metrics shift between stability classes, indicating limited universality of single-inference reliability when architecture or model version changes. 
This pattern is global and repeatable: it is not limited to a single pipeline or a single paired model comparison, but rather occurs consistently across all analysed comparisons.

It is important to emphasize that these visualizations are not intended to identify individual metrics, but rather to assess the distribution of reliability classes and their correspondence between models. In this sense, the concentration of points off the diagonal of the plots provides an empirical indicator of the sensitivity of metric stability to model choice, while the few points located on the diagonal represent metrics that maintain their reliability class regardless of the architecture used. Illustrative examples of such behaviour, relating to specific metrics and their stability classes, are presented in the summary tables and the Appendix.

These results indicate that high single-inference reliability is not a universal property for most metrics, but rather a characteristic of a limited subset of indicators. At the same time, the observed discrepancy in reliability classes between models highlights the importance of both the informed selection of metrics for a specific use case and the design of inference aggregation and replication strategies in multi-model systems.

Table \ref{tab:icc_number of metrics} provides a global summary of the distribution of metric reliability levels across models and reliability estimators. Unlike class- or pipeline-specific tables, this aggregation highlights the overall shift in reliability structure when moving from single-inference estimates (ICC(3,1)) to aggregated estimates (ICC(3,k)).

Across all models, the transition to ICC(3,k) results in a pronounced redistribution of metrics toward higher reliability classes, with a substantial increase in the proportion of metrics classified as excellent or perfect. At the same time, Table \ref{tab:icc_number of metrics} makes explicit that a non-trivial share of metrics remains in moderate or poor reliability classes at the level of ICC(3,1), even for the most stable model variants.

In this sense, Table \ref{tab:icc_number of metrics} serves as a reference point for interpreting subsequent analyses by separating the upper bound of achievable stability under aggregation from the baseline repeatability of single-pass inference. This distinction is critical for understanding which reliability gains are attributable to aggregation strategies rather than to inherent metric stability.

\begin{table}[ht]
\small
\centering
\small
\caption{Number of stable metrics (RT-consistent) according to class and models}
\label{tab:icc_number of metrics}
\resizebox{\textwidth}{!}{%
\begin{tabular}{lrrrrrrr}
\toprule
\textbf{Metric class} 
    & \textbf{GPT-4o audio} & \textbf{Gemini 2.0 Flash} & \textbf{Gemini 2.5 Flash}
    & \textbf{GPT-4o audio} & \textbf{Gemini 2.0 Flash} & \textbf{Gemini 2.5 Flash}
    & \textbf{Total} \\
    & \textbf{ICC(3,1)} & \textbf{ICC(3,1)} & \textbf{ICC(3,1)}
    & \textbf{ICC(3,k)} & \textbf{ICC(3,k)} & \textbf{ICC(3,k)}
    &  \\
\midrule
Adaptive                 & 2 &   0 &  10 &  11 &  3 &  11 &  11 \\
Affect alignment         & 8 &  3 &  4 & 28 & 22 & 23 & 64 \\
Cognitive style          & 14 & 0 & 14 & 23 & 3 & 14 & 23 \\
Engagement               & 24 & 17 & 44 & 40 & 26 & 45 & 46 \\
Intention                & 5 & 6 & 8 & 7 & 7 & 8 & 9 \\
Interactional efficiency & 6 & 0 & 14 & 12 & 1 & 14 & 15 \\
Personalization          & 12 & 11 & 4 & 12 & 12 & 11 & 12 \\
Relational synchrony     & 9 & 6 & 11 & 12 & 11 & 12 & 15 \\
Safety                   & 0 & 0 & 2 & 2 & 2 & 3 & 3 \\
Semantic appropriateness & 8 & 3 & 15 & 12 & 11 & 15 & 15 \\
\midrule
\textbf{Total per model} 
                        & 88 & 46 & 126 & 159 & 98 & 156 & 213 \\
\textbf{Share (\%)}    
                        & 41\% & 22\% & 59\% & 75\% & 46\% & 73\% &  100\% \\
\bottomrule
\end{tabular}
}%
\end{table}

\begin{table}[ht]
\centering
\small
\caption{Distribution of Metrics Interpretation for GPT-4o audio, Gemini 2.0 Flash and Gemini 2.5 Flash for two classes of coefficients (ICC(3) and ICC(3,k)}
\label{Distribution of metrics}
\resizebox{\textwidth}{!}{%
\begin{tabular}{lrrrrrr}
\toprule
\textbf{Reliability level} 
    & \textbf{GPT-4o audio} & \textbf{GPT-4o audio} 
    & \textbf{Gemini 2.0 Flash} & \textbf{Gemini 2.0 Flash} 
    & \textbf{Gemini 2.5 Flash} & \textbf{Gemini 2.5 Flash} \\
    & \textbf{ICC(3,1)} & \textbf{ICC(3,k)}
    & \textbf{ICC(3,1)} & \textbf{ICC(3,k)}
    & \textbf{ICC(3,1)} & \textbf{ICC(3,k)} \\
\midrule
\rowcolor[HTML]{90EE90}Perfect reliability 
    &   4 (2\%) &   0 &   56 (26\%) &   4 (2\%) &  0 &  56 (26\%) \\
\rowcolor[HTML]{90EE90}Excellent reliability 
    &  84 (39\%) & 46 (22\%) &  70 (33\%)&  155 (73\%) &  97 (46\%) & 100 (47\%) \\
\rowcolor[HTML]{E0FFE0}Good reliability 
    &  63 (30\%) &  43 (20\%) &  22 (10\%) &  24 (11\%) &  60 (28\%) &   9 (4\%) \\
\rowcolor[HTML]{FFFFE0}Moderate reliability 
    &  26 (12\%) &   48 (23\%) &  15 (7\%) &  9 (4\%) &  21 (10\%) &  10 (5\%) \\
\rowcolor[HTML]{FFE0E0}Poor reliability 
    &  36 (17\%) &   76 (36\%) &  50 (23\%) &  21 (10\%) &  34 (16\%) &  38 (18\%) \\
\midrule
\textbf{Total} 
    & 213 & 213 & 213 & 213 & 213 & 213 \\
\bottomrule
\end{tabular}
}%
\end{table}

The consistency of reliability classes between models was assessed using a common set of 213 metrics calculated for all three models analysed. The distribution of metric interpretation (from poor class to perfect reliability class) is presented in  Table \ref{Distribution of metrics}. 

For each metric, its reliability class was determined based on the ICC(3,1) value in each model. Then, for each metric and for each model, the assigned reliability class was compared with the classes obtained in the other models across three possible model pair comparisons (Gemini 2.0 Flash–Gemini 2.5 Flash, Gemini 2.0 Flash–GPT-4o audio, Gemini 2.5 Flash–GPT-4o audio). As a result, each metric was assigned a value from 0 to 3, corresponding to the number of model pairs in which the same ICC(3,1) reliability class was maintained.

A value of 0 indicates no consistency in the reliability class for any of the model pair comparisons, values of 1 and 2 indicate partial consistency, and a value of 3 indicates complete consistency in the reliability class for this metric across all three model pair comparisons. The obtained distributions within the received classes of consistency (0, 1, 2, 3) for each model are presented in Table \ref{tab: Consistency gemini 2.0}, Table \ref{Consistency gemini 2.5}, and Table \ref{Consistency gpt}.

\begin{table}[ht!]
\small
\centering
\caption{Gemini 2.0 Flash — Consistency Between Models for ICC(3,1) and ICC(3,k).
comparison results expressed as the number of concordant pairs between 3 models - 0 - no compatible pairs, 1 - one compatible pair, 2 - two compatible pairs, 3 - three compatible pairs}
\label{tab: Consistency gemini 2.0}
\begin{tabular}{lrrrrr|rrrrr}
\toprule
 & & ICC&(3,1) & & & & ICC&(3,k) & & \\
\midrule 
\textbf{Metric interpretation} 
    & {0} 
    & {1of3} 
    & {2of3} 
    & {3of3} 
    & {Total} 
    & {0} 
    & {1of3} 
    & {2of3} 
    & {3of3} 
    & {Total} \\
\midrule
\rowcolor[HTML]{90EE90}Excellent reliability      & 0  & 11 & 4  & 31 & 46 & 1 & 7 & 1 & 89 & 98 \\
\rowcolor[HTML]{E0FFE0}Good reliability           & 6  & 33 & 1  & 3  & 43 & 15 & 43 & 1 & 1 & 60 \\
\rowcolor[HTML]{FFFFE0}Moderate reliability       & 25 & 23 & 0  & 0  & 48 & 3 & 18 & 0 & 0 &21  \\
\rowcolor[HTML]{FFE0E0}Poor reliability           & 22 & 27 & 1  & 26 & 76 & 2& 17& 3& 12& 34\\
\midrule
\textbf{Total}             & 53 & 94 & 6  & 60 & 213 &21 & 85& 5& 102&213 \\
\bottomrule
\end{tabular}
\end{table}

\begin{table}[ht!]
\small
\centering
\label{Consistency gemini 2.5}
\caption{Gemini 2.5 Flash — Consistency Between Models for ICC(3,1) and ICC(3,k)
comparison results expressed as the number of concordant pairs between 3 models - 0 - no compatible pairs, 1 - one compatible pair, 2 - two compatible pairs, 3 - three compatible pairs}
\label{Consistency gemini 2.5}
\begin{tabular}{lrrrrr|rrrrr}
\toprule
 & & ICC&(3,1) & & & & ICC&(3,k) & & \\
\midrule 
\textbf{Metric interpretation} 
    & {0} 
    & {1of3} 
    & {2of3} 
    & {3of3} 
    & {Total} 
    & {0} 
    & {1of3} 
    & {2of3} 
    & {3of3} 
    & {Total} \\
\midrule
\rowcolor[HTML]{90EE90}Excellent reliability      & 39 & 52 & 4  & 31 & 126 & 6 & 60  & 1 & 89 & 156 \\
\rowcolor[HTML]{E0FFE0}Good reliability           & 1  & 17 & 1  & 3  & 22 & 1 & 6 & 1 & 1 & 9 \\
\rowcolor[HTML]{FFFFE0}Moderate reliability       & 3  & 12 & 0  & 0  & 15 & 2 & 8 & 0 & 0 & 10 \\
\rowcolor[HTML]{FFE0E0}Poor reliability           & 10 & 13 & 1  & 26 & 50 & 12 & 11 & 3 & 12 & 38 \\
\midrule
\textbf{Total}             & 53 & 94 & 6  & 60 & 213 & 21 & 85 & 5 & 102 & 213 \\
\bottomrule
\end{tabular}
\end{table}

\begin{table}[ht!]
\small
\centering
\label{Consistency gpt}
\caption{GPT-4o audio — Consistency Between Models for ICC(3,1) and ICC(3,k)
comparison results expressed as the number of concordant pairs between 3 models - 0 - no compatible pairs, 1 - one compatible pair, 2 - two compatible pairs, 3 - three compatible pairs}
\label{Consistency gpt}
\begin{tabular}{lrrrrr|rrrrr}
\toprule
 & & ICC&(3,1) & & & & ICC&(3,k) & & \\
\midrule 
\textbf{Metric interpretation} 
    & {0} 
    & {1of3} 
    & {2of3} 
    & {3of3} 
    & {Total} 
    & {0} 
    & {1of3} 
    & {2of3} 
    & {3of3} 
    & {Total} \\
\midrule
\rowcolor[HTML]{90EE90}Excellent reliability      & 8 & 45  & 4 & 31  & 88 & 6& 60&1 &89 & 156\\
\rowcolor[HTML]{E0FFE0}Good reliability           & 35 & 24 & 1  & 3  & 63 &1 &6 & 1& 1&9\\
\rowcolor[HTML]{FFFFE0}Moderate reliability       & 6  & 20 & 0  & 0 & 26 & 2& 8&0 & 0& 10\\
\rowcolor[HTML]{FFE0E0}Poor reliability           & 4  & 5  & 1 & 26 & 36 & 12& 11&3& 3& 38 \\
\midrule
\textbf{Total}             & 53 & 94 & 6  & 60 & 213 & 21 & 85 & 5 & 102 & 213\\
\bottomrule
\end{tabular}
\end{table}

Applying the described procedure shows that \textbf{only 31 of the 213 metrics analysed, or 14.6\%, achieve the excellent reliability class consistently in all three model pair comparisons}. This means that only a small subset of metrics maintains the highest ICC(3,1) reliability class regardless of the model choice.

The remaining high-reliability metrics in individual models demonstrate only partial or no consistency in the reliability class between models. \textbf{This indicates that the high stability of a single-inference is, in most cases, model-dependent and is not a universal property of the metric.}

The obtained quantitative result (14.6\%) constitutes an empirical measure of the limited universality of the reliability of a single inference in the tested set of metrics and sets an upper limit on the number of indicators that can be treated as fully comparable between different architectures and model versions without additional stabilization mechanisms.

\subsection{Study 3 results: Inter Consistency of Metric Values}
Study 3 extends the analysis to assess the consistency of the metric values between models. Because comparing metrics with low single-inference reliability would lead to conclusions dominated by random variance and measurement instability, this analysis was restricted to the subset of metrics that demonstrated full consistency of excellent reliability in Study 2: 31 for ICC(3,1) and 89 for ICC(3,k) in all three pairwise model comparisons. For this subset, we assess the degree of consistency of metric results between models, treating it as a test of measurement portability across architecture and model version changes. This consistency was captured by measures of ordinal correlation and/or absolute consistency and difference error, which allows us to distinguish truly model-agnostic metrics from metrics that are stable but systematically inconsistent across models.

\textbf{Method} The goal of the third stage of the study is to assess the consistency of metric values between language models, defined as the consistency of results generated by different models for the same units of analysis (conversation segments). This analysis is intentionally limited to metrics that meet the rigorous criteria of stability and universality defined in the two previous stages of the study. Only those metrics that demonstrated excellent reliability (for ICC(3,1) \& ICC(3,k) $\geq 0.9$) in all three models tested and maintained class consistency in all pairwise model comparisons were included in Study 3. This narrowing of the set of analysed metrics is a necessary methodological condition that enables a meaningful interpretation of the consistency of measurement values.

The unit of analysis in Study 3 is still a single conversation segment, for which each metric was calculated independently by each model. The same methodology was used as in Studies 1 and 2: for each combination (model × segment × metric), four independent model runs were performed, generating four measurement replicates. These replicates are not treated as independent observations but as replications of the same measurement, reflecting the generative randomness of the language models. Therefore, to avoid the arbitrary selection of a single run and the problem of pseudo-replication, the inter-model agreement analysis included all possible pairs of replicates between the compared models. For each model pair, agreement measures were calculated for all 16 combinations of pairs of runs (4 × 4), allowing for the generative variability (Studies 1 and 2) to be accounted for in the estimation of measurement comparability.

Table \ref{tab:study3_agreement} presents a summary of combined normalized Median Mean Absolute Error and median Cohen's Kappa agreement levels between model pairs: A - GPT-4o audio vs Gemini 2.0 Flash; B - GPT-4o audio vs Gemini 2.5 Flash; C - Gemini 2.0 Flash vs Gemini 2.5 across 31 features that demonstrated excellent single-inference reliability (ICC(3,1) $\geq 0.9$) and 89 features with excellent averaged-inferences reliability (ICC(3,k) $\geq 0.9$) in all three models. Cell values indicate count of features meeting each interpretation criterion. The A-3P (Agreement across 3 Pairs) column represents the count of features where all three model pairs (A, B, and C) fell into the same interpretation category simultaneously.  

\begin{table}[]
\centering
\caption{Summary of combined normalized Median Mean Absolute Error and median Cohen's Kappa agreement levels between model pairs:  A - GPT-4o audio vs Gemini 2.0 Flash; B - GPT-4o audio vs Gemini 2.5 Flash; C - Gemini 2.0 Flash vs Gemini 2.5.
A-3P column is agreement across 3 pairs)}

\label{tab:study3_agreement}
\begin{tabular}{lllllllll}
\hline
\multirow{2}{*}{\textbf{\begin{tabular}[c]{@{}l@{}}Interpretation\\ of agreement\end{tabular}}} &
  \multicolumn{4}{l}{\textbf{ICC(3,1)}} &
  \multicolumn{4}{l}{\textbf{ICC(3,k)}} \\ \cline{2-9} 
 &
  \textbf{A} &
  \textbf{B} &
  \textbf{C} &
  \textbf{A-3P} &
  \textbf{A} &
  \textbf{B} &
  \textbf{C} &
  \textbf{A-3P} \\ \hline
\rowcolor[HTML]{90EE90}near-ideal     & 19 & 9  & 9  & 9  & 35 & 22 & 28 & 19 \\
\rowcolor[HTML]{E0FFE0}moderate       & 4  & 6  & 5  & 1  & 21 & 22 & 17 & 7  \\
\rowcolor[HTML]{FFFFE0}low            & 4  & 7  & 10 & 1  & 11 & 5  & 7  & 6  \\
\rowcolor[HTML]{FFE0E0}non-acceptable & 4  & 9  & 7  & 5  & 15 & 24 & 16 & 5  \\ \hline
Total          & 31 & 31 & 31 & 16 & 89 & 89 & 89 & 37 \\ \hline
\end{tabular}
\end{table}

Detailed measure selection is reported at the metric level. For each metric, the set of 16 agreement values for each model pair is statistically summarized. The median of the concordance distribution was used as the primary outcome measure. This represents a conservative estimate of the typical level of measurement comparability and is robust to single, extreme deviations resulting from generative randomness (if they occurred despite the selection of the most stable metrics). 
Additionally, to assess the sensitivity of the results to the choice of aggregation measure, the trimmed mean was reported as a supplementary analysis. This approach preserves information about the central part of the concordance distribution while limiting the influence of outliers.

This procedure allows for the assessment of the concordance of metric values between models in a way that does not rely on the arbitrary selection of a single measurement or the assumption of the independence of replicates. Treating replicates as replicates and using robust summary statistics allows for the separation of systematic differences between models from random generative variability. Consequently, the obtained concordance measures reflect true measurement comparability, not artifacts resulting from individual model implementations.

For metrics evaluated under the ICC(3,1) criterion, 31 out of 213 metrics met the inclusion criteria. Among these, only 9 metrics achieved near-ideal agreement simultaneously across all three model pairs (A–3P), indicating limited cross-model comparability at the level of single inferences.

For metrics evaluated under ICC(3,k), aggregation substantially increased the number of metrics meeting the stability and universality criteria, with 89 out of 213 metrics qualifying for inter-model agreement analysis. However, this increase in stable metrics did not translate into proportional agreement of metric values across models. Only 19 of the 89 metrics achieved near-ideal agreement across all three model pairs.

These results (Table \ref{tab:study3_agreement} show that while aggregation improves measurement stability and expands the pool of metrics eligible for comparison, stability alone does not guarantee agreement of metric values between models. Consequently, improved reliability through aggregation should not be interpreted as evidence of measurement equivalence or validity in cross-model settings.

\section{Interpretation}

The three-stage validation strategy adopted in this study yields three distinct, yet complementary, types of knowledge about the measurement of user states using LLMs. Each study addresses a different analytical question and supports a different level of inference. Making this distinction explicit is essential in order to avoid conflating measurement properties with substantive interpretations of model behaviour.

At the first level, Study 1 addresses the most fundamental question: \textbf{does a given configuration (model, analytical pipeline, and interaction context) produces measurements that are sufficiently stable to be considered meaningful?} The goal is to determine whether repeated assessments of identical input conditions yield consistent metric values that can justify interpretation at the level of individual outputs. At this level, stability is treated as a prerequisite for interpreting any single metric value as an indicator of the user’s current state. 

At the second level, Study 2 examines whether the reliability classes assigned in Study 1 are invariant across models. The focus is not on the numerical agreement of metric values, but on \textbf{whether the same metric retains its reliability status independently of the model used for evaluation}. This level of analysis tests the assumption that measurement reliability is a property of the metric itself rather than an artifact of a specific model configuration. 

The third level of analysis, developed in Study 3, concerns \textbf{the comparability of inferred metric values between models} treating different models as alternative measurement instruments rather than as validated external criteria. This analysis is intentionally restricted to metrics that satisfy the stability and universality conditions established in the preceding studies. The objective is to assess whether such metrics exhibit high inter-model agreement in their values when applied to identical input data, thereby supporting meaningful cross-model comparison. 

Taken together, the results indicate that metric reliability is configuration-dependent, rarely universal, and progressively degrades as increasingly strict interpretability conditions are imposed. \textbf{Stability cannot be treated as a default characteristic of user state metrics inferred by LLMs, but must instead be empirically established for each intended mode of use.}
This provides a cumulative assessment of the assumptions formulated in Hypotheses H1–H3. Rather than being tested in isolation, these hypotheses are progressively constrained as additional stability conditions are introduced, moving from within-model repeatability, through inter-model robustness, to cross-model agreement of metric values.
This layered design reveals that optimistic assumptions regarding metric stability are increasingly difficult to sustain as the requirements for interpretability become more stringent. In this sense, \textbf{the findings do not invalidate individual hypotheses independently (there are some metrics which fulfil criteria), but collectively demonstrate that reliability, universality, and resistance to inter-model variability cannot be jointly assumed for most user state metrics.}
From the perspective of HAI, these findings imply that adaptive systems relying on inferred user states cannot assume stable or transferable measurements across models or deployments. Metrics that may be suitable for global or post-hoc analyses do not necessarily meet the requirements for real-time adaptive use. Employing unstable or model-specific metrics as direct adaptation signals risks introducing inconsistent or unpredictable system behaviour driven by inferential variability rather than genuine changes in user state.

The implications are particularly salient for the design of AI agents and agentic systems based on large language models. Agents that condition their planning, decision-making, or goal adaptation on inferred user states implicitly assume that these signals are stable and comparable across model versions and deployment contexts. 
The present findings show that this assumption is generally unwarranted. \textbf{Reliability-validated metrics should therefore be treated as part of the agent’s cognitive grounding, while metrics that do not meet stability requirements must be calibrated or excluded from autonomous decision loops. In this sense, reliability is not an auxiliary evaluation criterion but a foundational condition for responsible agentic behaviour.}
  
\section{Discussion}
Hypotheses 1 and 2 were formulated to reflect commonly held expectations regarding the stability of user state metrics across applications. However, the empirical results do not support these expectations and reveal significant limitations to their reliability. The research confirms Hypothesis 3, which states that metric reliability varies depending on the model used, but the magnitude of this variation is surprisingly large.
Although the results do not support a binary interpretation of the hypotheses, they reveal a structured pattern of stability degradation: while a subset of metrics exhibits acceptable stability within a single model and under aggregation, stability deteriorates substantially at the level of real-time use and becomes highly model-dependent in cross-model settings.
Thus, the reliability of user state inferences generated by LLMs cannot be assumed. This was verified under rigorous conditions: constant input data and unchanged model configuration. This means that inferences about user states made by LLMs do not automatically meet psychometric measurement criteria, and their interpretability is conditional and strictly dependent on stability.

\textbf{Reliability, understood as the stability of inference, is a condition for explainability. A key result of this work is the distinction between the semantic meaningfulness of responses and measurement reliability. Models can generate convincing responses at the content level while simultaneously lacking repeatability in replications. However, such variability is not interpretable in a manner analogous to the variability observed in humans. In the case of human assessments (regardless of whether it is self-assessment or the assessment of others), inconsistency does not necessarily indicate error and can be informative about the internal state, perspective, context, or will of the assessor. In the case of AI, the situation is fundamentally different. Models lack internal states or intentionality that would make sense of discrepancies between successive inferences. The variability of AI results does not convey information but rather indicates the instability of the inferential process, which affects expectations. Therefore, reliability is a necessary condition for explainability, as without it, it is impossible to analyse causes or construct a coherent interpretation of results.}

The results of the conducted research do not justify assessing the "quality" of models or ranking them in terms of their ability to recognize user states. The study does not specifically address accuracy relative to the supposed ground truth, as humans do not constitute a stable reference point in interpretive domains. Therefore, only the internal consistency and repeatability of AI inferences under controlled, and laboratory conditions of repeatability are analysed.

Psychological constructs themselves are not questioned here. The only assumption that the inferential tasks assigned to LLMs enable their reliable measurement at the level of individual inferences, especially in real-time adaptive applications, is questioned. The observed increase in measured metric stability after aggregation should not be interpreted as a "fix" for unstable inferences. Aggregation merely changes the unit of analysis and shifts the interpretation from the state at a given moment to the level of the model's average behavior over time. Consequently, the stability achieved through aggregation cannot retrospectively justify decisions made based on individual inferences.

An important implication of the results is that reliability is not a permanent property. Even metrics that meet the stability criteria at a given moment can lose this stability due to changes in context, input data, or implicit drift in model behaviour over time. Therefore, all applications of user state inference using LLMs require continuous reliability monitoring rather than a one-time validation.

In this context, it is reasonable to consider early warning mechanisms, whose role is not to improve or adapt models but rather to signal the point at which the metrics used for evaluation no longer meet the minimum requirements for interpretability and explainability.


\section{Summary \& Conclusions}
The aim of this work was to examine the reliability of metrics used to assess user states using AI. Measurement reliability, following the principles of psychometrics, is understood as a necessary condition for the valid interpretation of these measurement results in various application contexts. The study does not address validity with respect to presumed ground truth or comparisons of model quality, but focuses on the repeatability of metrics under replication conditions as a minimum laboratory requirement for their use in adaptive and analytical systems.

The results indicate that, in interpretive domains, metric reliability definitely cannot be considered a default property, and that the absence of reliability decisively precludes the interpretation of individual ratings as stable indicators of user state. This implies that the semantic meaningfulness of results generated by language models is not a sufficient basis for treating them as measurement signals, especially in real-time applications.

At the same time, the study demonstrates that a lack of reliability at the level of individual ratings does not automatically invalidate the analytical value of a metric. Metrics that are unstable individually can retain their utility in aggregated analyses, identifying rules governing interaction patterns and the relationships between system characteristics and variables such as satisfaction, trust, or user engagement. However, it is crucial to clearly separate these two classes of applications, which, in practice, can be inappropriately mixed.

A significant contribution of this work is proposing an evaluation framework that allows for empirically distinguishing the usefulness of metrics for real-time adaptation from their usefulness for post-hoc analytics. This distinction is methodological in nature and independent of a specific model, domain, or dataset, allowing it to be replicated and applied in other studies on AI-based user state assessment.

From a practical perspective, the results indicate that using metrics without prior validation of their reliability leads to spurious adaptations based on unstable signals that cannot be meaningfully interpreted or explained. Consequently, any pipelines using AI to assess user states require not only initial validation but also ongoing reliability monitoring over time. 
More broadly, this work contributes to the discussion on responsible AI use, demonstrating that the issue of metric reliability is not merely a technical issue, but a prerequisite for correctly interpreting results in systems that influence user experience and behaviour. The proposed approach does not offer simple optimization solutions, but rather provides a threshold criterion for determining when AI-generated assessments can be used as measurement signals and when they should remain solely a tool for descriptive analysis.

\printbibliography
\appendix
\input{icc_comparison_table}
\input{feature_prompts_table} 

\input{full_study_3_tables}

\end{document}

\subsubsection{Basic Statistics}
\begin{itemize}
    \item \textbf{Total Conversations}: 15
    \item \textbf{Total Duration}: 52 minutes 15 seconds (3,136 seconds)
    \item \textbf{Average Conversation Length}: 3.5 minutes (209 seconds)
    \item \textbf{Total Audio Segments}: 552
    \item \textbf{Average Segments per Conversation}: 36.8
\end{itemize}

\subsubsection{Duration Distribution}
\begin{itemize}
    \item \textbf{Shortest Conversation}: 187.4 seconds ($\sim$3.1 minutes)
    \item \textbf{Longest Conversation}: 234.0 seconds ($\sim$3.9 minutes)
    \item \textbf{Median Duration}: 208.0 seconds (3.5 minutes)
\end{itemize}

\subsection{Content Analysis}

\begin{table}[h!]
\centering
\begin{tabular}{@{}lcc@{}}
\toprule
\textbf{Topic Category} & \textbf{Count} & \textbf{Percentage} \\
\midrule
Product Information/Sales & 6 & 13.6\% \\
Technical Support & 5 & 11.4\% \\
Account Management & 5 & 11.4\% \\
Service Activation/Deactivation & 5 & 11.4\% \\
Equipment/Device Support & 4 & 9.1\% \\
Complaints and Issues & 4 & 9.1\% \\
Contract/Subscription Changes & 4 & 9.1\% \\
Billing and Payments & 4 & 9.1\% \\
\bottomrule
\end{tabular}
\caption{Topic Distribution in the Dataset}
\label{tab:topics}
\end{table}

The dataset covers diverse telecommunications customer service scenarios that are covered in Table \ref{tab:topics}. Altogether, the dataset includes 30 unique speakers, with a clear predominance of female voices (63\%, see Table~\ref{tab:gender_overall}). The roles of customers and consultants are evenly split (15 each), but their gender distributions differ noticeably. 

Among the customers, the balance is approximately even. 
\begin{itemize}
    \item \textbf{Male}: 8 speakers (53.3\%)
    \item \textbf{Female}: 7 speakers (46.7\%)
\end{itemize}

In contrast, the consultant group is predominantly female: 
\begin{itemize}
    \item \textbf{Female}: 12 speakers (80.0\%)
    \item \textbf{Male}: 3 speakers (20.0\%)
\end{itemize}

This asymmetry may reflect the staffing patterns often found in customer support services.

\begin{table}[h!]
\centering
\begin{tabular}{@{}lcc@{}}
\toprule
\textbf{Gender} & \textbf{Count} & \textbf{Percentage} \\
\midrule
Female & 19 & 63.3\% \\
Male & 11 & 36.7\% \\
\bottomrule
\end{tabular}
\caption{Overall Gender Distribution}
\label{tab:gender_overall}
\end{table}

\begin{longtable}{@{}p{0.5\textwidth}p{0.4\textwidth}@{}}
\toprule
\textbf{Metric} & \textbf{Value} \\
\midrule
\endhead
Dataset Size & 15 conversations \\
Total Duration & 52 minutes 15 seconds \\
Average Conversation Length & 3.5 minutes \\
Total Audio Segments & 552 \\
Segments per Conversation (avg) & 36.8 \\
Primary Language & Polish \\
Domain & Telecommunications Customer Service \\
Most Common Topic & Product Information/Sales \\
Female Speakers & 19 (63.3\%) \\
Male Speakers & 11 (36.7\%) \\
Audio Quality & Phone recordings \\
Transcription Quality & High (manual transcription) \\
\bottomrule
\caption{Summary of Key Dataset Statistics}
\label{tab:key_stats}
\end{longtable}

\begin{table}
\centering
\caption{Reliability by metric class: min, max, mean ICC and counts of `excellence' (\(\geq 0.90\)) and `good' (\([0.75, 0.90)\)) metrics for ICC(3,1) and ICC(3,k).}
\label{tab:icc_class_summary}
\begin{tabular}{lrrrrrrrrrr}
\toprule
pipeline\_type & ICC3,1\_min & ICC3,1\_max & ICC3,1\_mean & ICC3,1\_excellence\_n & ICC3,1\_good\_n & ICC3,k\_min & ICC3,k\_max & ICC3,k\_mean & ICC3,k\_excellence\_n & ICC3,k\_good\_n \\
\midrule
adaptive & 0.381 & 0.846 & 0.560 & 0 & 2 & 0.711 & 0.957 & 0.823 & 2 & 2 \\
affect\_alignment & -0.019 & 0.986 & 0.483 & 12 & 12 & -0.081 & 0.996 & 0.671 & 23 & 22 \\
cognitive\_style & 0.096 & 0.744 & 0.498 & 0 & 0 & 0.298 & 0.921 & 0.768 & 3 & 3 \\
engagement & 0.003 & 0.998 & 0.698 & 13 & 19 & 0.013 & 0.999 & 0.810 & 11 & 26 \\
intention & 0.554 & 0.967 & 0.860 & 5 & 3 & 0.833 & 0.991 & 0.954 & 2 & 6 \\
interactional\_efficiency & 0.393 & 0.727 & 0.562 & 0 & 0 & 0.678 & 0.914 & 0.817 & 3 & 1 \\
personalization & 0.959 & 0.999 & 0.991 & 10 & 2 & 0.999 & 1.000 & 0.999 & 12 & 0 \\
relational\_synchrony & 0.667 & 0.994 & 0.889 & 6 & 3 & 0.992 & 0.999 & 0.963 & 12 & 0 \\
safety & 0.824 & 0.977 & 0.927 & 4 & 0 & 0.975 & 0.994 & 0.970 & 3 & 0 \\
semantic\_appropriateness & 0.303 & 0.924 & 0.726 & 3 & 2 & 0.826 & 0.980 & 0.934 & 11 & 1 \\
\bottomrule
\end{tabular}
\end{table}

\textbf{Intraclass Correlation Coefficient (ICC)}

As the most appropriate reliability indicator for repeated continuous measurements of the same stimulus, we used the Intraclass Correlation Coefficient (ICC), specifically:

\begin{itemize}
    \item \textbf{ICC(3,1)}: reliability of a single measurement under a two-way mixed-effects model (fixed raters), assessing absolute consistency across repetitions.
    \item \textbf{ICC(3,k)}: reliability of the mean of $k$ repeated measurements for each segment.
\end{itemize}

The formula for \textbf{ICC(3,1)} is:

\[
\text{ICC}(3,1) = \frac{MS_B - MS_W}{MS_B + (k - 1)MS_W}
\]

The formula for \textbf{ICC(3,k)} is:

\[
\text{ICC}(3,k) = \frac{MS_B - MS_W}{MS_B}
\]

where:
\begin{itemize}
    \item $MS_B$ is the mean square between targets (segments),
    \item $MS_W$ is the mean square error (within-segment variance across repetitions),
    \item $k$ is the number of repetitions (e.g., $k=4$).
\end{itemize}

These indicators reflect the proportion of total variance attributable to differences between segments rather than measurement noise.

Commonly used reliability indices such as Pearson or Spearman correlation coefficients, or internal consistency measures like Cronbach’s alpha or McDonald’s omega, were deemed inappropriate in this context. Correlations capture only covariation, not absolute agreement, and are sensitive to systematic shifts \cite{brzezinski1996metodologia}. Alpha and omega assume independence of items and the presence of a latent trait, conditions violated in our setup where repetitions are technical replications, not itemized tests \cite{magnusson1971test}.

CC(3,1) and ICC(3,k) are the most valid and interpretable tools for assessing the test--retest reliability of automated user state classification systems under fixed conditions. Their application enables the quantification of model-level measurement stability, which is a prerequisite for trustworthy AI-based inference.

\paragraph{Affect Alignment.} Recognize the user's emotional state and adjust tone/style to reduce negative affect and increase relational safety. Typical signals: valence--arousal indices, prosodic features (tempo, intensity, pitch variation), affective lexicons; levers: apologies, empathic paraphrase, de-escalation. \textit{Use when complaints, crises, or heightened arousal are detected.}

\paragraph{Engagement.} Maintain active participation and attention. Signals: turns/min, utterance length, response latency, open questions by the user, trend in activity (\(\Delta\)~engagement). Levers: topic switches, curiosity prompts, humour, difficulty scaffolding. \textit{High weight in sales/education; lower in task resolution.}

\paragraph{Cognitive Fit.} Match linguistic complexity and register to the user's cognitive style. Signals: readability (e.g., FKRS), syntactic depth, concreteness/abstraction, domain terminology density. Levers: simplification vs. elaboration, analogy vs. concrete examples, register shift. \textit{Key in technical support and tutoring.} 

\paragraph{Relational Synchrony.} Align formalities, politeness strategies, humour, and tonal stance. Signals: formality detectors, politeness markers, backchannels/acknowledgements, tonal entrainment. Levers: raising/lowering formality, increasing acknowledgements, switching stance (expert vs.~supportive). \textit{Critical for trust and long-term retention.}

\paragraph{Semantic Appropriateness.} Ensure answers are on-topic, precise, and coherent with dialog context. Signals: semantic similarity (e.g., SBERT cosine), keyword overlap, topical coherence/entropy, self-repair rate. Levers: clarification questions, paraphrase, topic refocus. \textit{Essential for factual trust.}

\paragraph{Personalization (Personality-informed Tailoring).} Adapt content and style to stable user traits/preferences and conversational history. Signals: personality cues (e.g., OCEAN from text), topic/stylistic adherence, history-informed references. Levers: register/tempo/topic adaptation, persuasion style alignment. \textit{High impact on satisfaction and return use.} 

\paragraph{Predictability \& Safety.} Control semantic, emotional, legal, and cognitive risks. Signals: risk lexicons, ambiguity/entropy scores, hallucination detectors, off-topic rate. Levers: refusal/deferral, hand-off, simplification, constrained generation. \textit{Mandatory in regulated/sensitive domains.}

\paragraph{Adaptive Timing.} Decide \emph{when} to switch strategy (tone/topic/register). Signals: lack of progress (loops), drops in relevance or engagement, sudden affect shifts. Levers: strategy reset, goal reframing, pacing changes. \textit{Increases naturalness and task progress.}

\paragraph{Goal Alignment.} Establish, track, and maintain the user's goal; avoid intent drift. Signals: explicit goal mentions, intent--response congruence, re-goaling counts, time-to-goal. Levers: paraphrase-and-confirm, targeted disambiguation, decision shortcuts. \textit{Core for task success.}

\paragraph{Interactional Efficiency.} Achieve outcomes with minimal communicative overhead. Signals: turns-to-goal, time-to-goal, repetition/paraphrase counts, brevity vs.~informativeness balance, off-topic rate. Levers: compression, step merging, guidance to action. \textit{Primary in transactional/helpdesk flows.}

\begin{table}[ht!]
\small
\centering
\caption{Gemini 2.0 Flash — Consistency Between Models for ICC(3,1).
comparison results expressed as the number of concordant pairs between 3 models - 0 - no compatible pairs, 1 - one compatible pair, 2 - two compatible pairs, 3 - three compatible pairs}
\label{tab: Consistency gemini 2.0}
\begin{tabular}{lrrrrr}
\toprule
\textbf{Metric interpretation} 
    & \textbf{0} 
    & \textbf{1 of 3} 
    & \textbf{2 of 3} 
    & \textbf{3 of 3} 
    & \textbf{Total} \\
\midrule
\rowcolor[HTML]{90EE90}Excellent reliability      & 0  & 11 & 4  & 31 & 46\,(22\%) \\
\rowcolor[HTML]{E0FFE0}Good reliability           & 6  & 33 & 1  & 3  & 43\,(20\%) \\
\rowcolor[HTML]{FFFFE0}Moderate reliability       & 25 & 23 & 0  & 0  & 48\,(23\%)  \\
\rowcolor[HTML]{FFE0E0}Poor reliability           & 22 & 27 & 1  & 26 & 76\,(36\%) \\
\midrule
\textbf{Total}             & 53 & 94 & 6  & 60 & 213\,(100\%) \\
\bottomrule
\end{tabular}
\end{table}

\begin{table}[ht!]
\small
\centering
\label{Consistency gemini 2.5}
\caption{Gemini 2.5 Flash — Consistency Between Models for ICC(3,1)
comparison results expressed as the number of concordant pairs between 3 models - 0 - no compatible pairs, 1 - one compatible pair, 2 - two compatible pairs, 3 - three compatible pairs}
\label{Consistency gemini 2.5}
\begin{tabular}{lrrrrr}
\toprule
\textbf{Metric interpretation} 
    & \textbf{0} 
    & \textbf{1 of 3} 
    & \textbf{2 of 3} 
    & \textbf{3 of 3} 
    & \textbf{Total} \\
\midrule
\rowcolor[HTML]{90EE90}Excellent reliability      & 39 & 52 & 4  & 31 & 126\,(59\%) \\
\rowcolor[HTML]{E0FFE0}Good reliability           & 1  & 17 & 1  & 3  & 22\,(10\%)  \\
\rowcolor[HTML]{FFFFE0}Moderate reliability       & 3  & 12 & 0  & 0  & 15\,(7\%)   \\
\rowcolor[HTML]{FFE0E0}Poor reliability           & 10  & 13 & 1  & 26 & 50\,(23\%)  \\
\midrule
\textbf{Total}             & 53 & 94 & 6  & 60 & 213\,(100\%) \\
\bottomrule
\end{tabular}
\end{table}

\begin{table}[ht!]
\small
\centering
\label{Consistency gpt}
\caption{GPT-4o audio — Consistency Between Models for ICC(3,1) 
comparison results expressed as the number of concordant pairs between 3 models - 0 - no compatible pairs, 1 - one compatible pair, 2 - two compatible pairs, 3 - three compatible pairs}
\label{Consistency gpt}
\begin{tabular}{lrrrrr}
\toprule
\textbf{Metric interpretation} 
    & \textbf{0} 
    & \textbf{1 of 3} 
    & \textbf{2 of 3} 
    & \textbf{3 of 3} 
    & \textbf{Total} \\
\midrule
\rowcolor[HTML]{90EE90}Excellent reliability      & 8 & 45  & 4 & 31  & 88\,(41\%) \\
\rowcolor[HTML]{E0FFE0}Good reliability           & 35 & 24 & 1  & 3  & 63\,(30\%) \\
\rowcolor[HTML]{FFFFE0}Moderate reliability       & 6  & 20 & 0  & 0 & 26\,(12\%) \\
\rowcolor[HTML]{FFE0E0}Poor reliability           & 4  & 5  & 1  & 26 & 36\,(17\%) \\
\midrule
\textbf{Total}             & 53 & 94 & 6  & 60 & 213\,(100\%) \\
\bottomrule
\end{tabular}
\end{table}

%% file: icc_comparison_table.tex
\definecolor{good}{HTML}{66CC66}      
\definecolor{excellent}{HTML}{009900} 
\definecolor{moderate}{HTML}{CCAA00}  
\definecolor{poor}{HTML}{CC6666}      
\definecolor{neg}{HTML}{B0C4DE}       

\scriptsize
\begin{landscape}
\begin{longtable}{llcccccc}
\caption{ICC3 and ICC3K values for all models across pipelines and metrics. 
Cell colours indicate reliability: 
\textcolor{excellent}{Excellent (0.90-1.0)}, 
\textcolor{good}{Good (0.75-$<$0.90)}, 
\textcolor{moderate}{Moderate (0.50-$<$0.75)}, 
\textcolor{poor}{Poor ($<$ 0.50).},
\textcolor{neg}{NaNs or $<$ 0.0}
}
\label{tab:icc_comparison}\\
\hline
\multicolumn{1}{c}{} &
  \multicolumn{1}{c}{} &
  \multicolumn{3}{c}{\textbf\small{ICC(3,1)}} &
  \multicolumn{3}{c}{\textbf\small{ICC(3,k)}} \\ \cline{3-8} 
\multicolumn{1}{c}{\multirow{-2}{*}{\textbf\small{Pipeline}}} &
  \multicolumn{1}{c}{\multirow{-2}{*}{\textbf\small{Metric}}} &
  \textbf\small{Gem 2.0} &
  \textbf\small{Gem 2.5} &
  \textbf\small{GPT-4o} &
  \textbf\small{Gem 2.0} &
  \textbf\small{Gem 2.5} &
  \textbf\small{GPT-4o} \\ \hline
\endfirsthead
\hline
\multicolumn{1}{c}{} &
  \multicolumn{1}{c}{} &
  \multicolumn{3}{c}{\textbf\small{ICC(3,1)}} &
  \multicolumn{3}{c}{\textbf\small{ICC(3,k)}} \\ \cline{3-8} 
\multicolumn{1}{c}{\multirow{-2}{*}{\textbf\small{Pipeline}}} &
  \multicolumn{1}{c}{\multirow{-2}{*}{\textbf\small{Metric}}} &
  \textbf\small{Gem 2.0} &
  \textbf\small{Gem 2.5} &
  \textbf\small{GPT-4o} &
  \textbf\small{Gem 2.0} &
  \textbf\small{Gem 2.5} &
  \textbf\small{GPT-4o} \\ \hline
\endhead
\hline
\endfoot
\endlastfoot
adaptive &
  actual\_strategy\_time.same\_strategy\_turns &
  \cellcolor[HTML]{FFE0E0}0,464 &
  \cellcolor[HTML]{90EE90}0,978 &
  \cellcolor[HTML]{E0FFE0}0,829 &
  \cellcolor[HTML]{E0FFE0}0,776 &
  \cellcolor[HTML]{90EE90}0,994 &
  \cellcolor[HTML]{90EE90}0,951 \\
adaptive &
  ambiguity\_resolving\_cost.ambiguity\_cost &
  \cellcolor[HTML]{FFE0E0}0,48 &
  \cellcolor[HTML]{90EE90}0,992 &
  \cellcolor[HTML]{E0FFE0}0,877 &
  \cellcolor[HTML]{E0FFE0}0,787 &
  \cellcolor[HTML]{90EE90}0,998 &
  \cellcolor[HTML]{90EE90}0,966 \\
adaptive &
  ambiguity\_resolving\_cost.clarification\_turns &
  \cellcolor[HTML]{FFE0E0}0,399 &
  \cellcolor[HTML]{90EE90}0,996 &
  \cellcolor[HTML]{E0FFE0}0,842 &
  \cellcolor[HTML]{FFFFE0}0,727 &
  \cellcolor[HTML]{90EE90}0,999 &
  \cellcolor[HTML]{90EE90}0,955 \\
adaptive &
  ambiguity\_resolving\_cost.is\_ambiguous\_query &
  \cellcolor[HTML]{FFE0E0}0,44 &
  \cellcolor[HTML]{90EE90}0,985 &
  \cellcolor[HTML]{E0FFE0}0,825 &
  \cellcolor[HTML]{E0FFE0}0,758 &
  \cellcolor[HTML]{90EE90}0,996 &
  \cellcolor[HTML]{90EE90}0,95 \\
adaptive &
  dialog\_progression.findings.loop\_detected &
  \cellcolor[HTML]{E0FFE0}0,779 &
  \cellcolor[HTML]{90EE90}0,99 &
  \cellcolor[HTML]{E0FFE0}0,801 &
  \cellcolor[HTML]{90EE90}0,934 &
  \cellcolor[HTML]{90EE90}0,997 &
  \cellcolor[HTML]{90EE90}0,942 \\
adaptive &
  dialog\_progression.stagnation\_score &
  \cellcolor[HTML]{E0FFE0}0,753 &
  \cellcolor[HTML]{90EE90}0,985 &
  \cellcolor[HTML]{E0FFE0}0,882 &
  \cellcolor[HTML]{90EE90}0,924 &
  \cellcolor[HTML]{90EE90}0,996 &
  \cellcolor[HTML]{90EE90}0,968 \\
adaptive &
  follow\_up\_query\_type.confidence &
  \cellcolor[HTML]{FFFFE0}0,605 &
  \cellcolor[HTML]{90EE90}0,983 &
  \cellcolor[HTML]{E0FFE0}0,89 &
  \cellcolor[HTML]{E0FFE0}0,859 &
  \cellcolor[HTML]{90EE90}0,996 &
  \cellcolor[HTML]{90EE90}0,97 \\
\hline
\end{longtable}
\end{landscape}

%% file: feature_prompts_table.tex



\newcolumntype{L}[1]{>{\RaggedRight\arraybackslash}p{#1}}

\scriptsize

\setlength{\tabcolsep}{6pt}
\renewcommand{\arraystretch}{1.8}

\begin{longtable}{|L{0.18\textwidth}|L{0.44\textwidth}|L{0.34\textwidth}|}
\caption{Complete Systematized Feature Descriptions and Data Structures} \label{tab:metrics_full} \\
\hline
\rowcolor{gray!20} \textbf{Metric Name} & \textbf{Description \& Formula Details} & \textbf{Dictionary Outputs} \\
\hline
\endfirsthead

\multicolumn{3}{c}{{\bfseries \tablename\ \thetable{} -- continued from previous page}} \\
\hline
\rowcolor{gray!20} \textbf{Metric Name} & \textbf{Description \& Formula Details} & \textbf{Dictionary Outputs} \\
\hline
\endhead

\hline \multicolumn{3}{|r|}{{Continued on next page}} \\ \hline
\endfoot

\hline
\endlastfoot

\hline \multicolumn{3}{|l|}{\cellcolor{gray!10}\textbf{Adaptive Dynamics Analysis}} \\ \hline

Actual Strategy Time & 
Measures the duration (consecutive turns or time units) a consultant maintains a single consistent communication strategy (e.g., empathetic listening). & 
\texttt{'same\_strategy\_turns'}: int \newline
\texttt{'strategy\_name'}: str \\ \hline

User Patterns Changing & 
Analyzes customer messages to detect shifts in linguistic complexity (0-100\% change in number of words) and emotional state (emotional valence from -1 to 1) based on linguistic and sentiment changes. & 
\texttt{'last\_user\_message'}: str \newline
\texttt{'emotional\_valence'}: float \newline
\texttt{'linguistic\_complexity'}: float \\ \hline

Tone Topic Fluctuations & 
Calculates the entropy score for the conversation window. \newline
$\texttt{transition\_entropy} = - \sum [P(strategy_i) \cdot \log_2(P(strategy_i))]$ \newline
\smallskip \textbf{Variables:} \newline
$P(strategy_i) = \text{Count}(strategy_i) / \text{TotalTurns}$. \newline
$\sum$: Sum over unique strategies in the window. & 
\texttt{'transition\_entropy\_score'}: float \newline
\texttt{'strategy\_distribution'}: dict\textless{}float\textgreater{} \newline
\texttt{'assessment'}: str \\ \hline

Response Accuracy Decrease & 
Determines if system relevance is decaying. \newline
$\texttt{relevance\_decay} = \text{slope}(\text{LinearRegression}([sem\_rel(t-n) \dots]))$ \newline
\smallskip \textbf{Variables:} \newline
$sem\_rel$: Cosine similarity between query and answer embeddings. \newline
$\text{slope} < 0$: Negative decay. & 
\texttt{'relevance\_decay'}: float \newline
\texttt{'trend'}: \textless{} Improving | Stable | Decaying \textgreater \\
\hline

\end{longtable}

%% file: full_study_3_tables.tex

\LTcapwidth=\textwidth
\begin{longtable}[c]{llllllll}
\caption{{Study 3 table with consistency of the feature values between the following model pairs: A - GPT4o vs Gemini 2.0; B - GPT4o vs Gemini 2.5; C - Gemini 2.0 vs Gemini 2.5 measured by normalized median Mean Absolute Error for continuous features and median Cohen's Kappa for binary features.}}
\label{tab:study3_full_89}\\
\hline
\multicolumn{1}{c}{} &
  \multicolumn{1}{c}{} &
  \multicolumn{3}{c}{\textbf{norm med MAE}} &
  \multicolumn{3}{c}{\textbf{med $\kappa$}} \\
\endfirsthead
\multicolumn{8}{c}%
{{\bfseries Table \thetable\ continued from previous page}} \\
\hline
\multicolumn{1}{c}{} &
  \multicolumn{1}{c}{} &
  \multicolumn{3}{c}{\textbf{norm med MAE}} &
  \multicolumn{3}{c}{\textbf{med $\kappa$}} \\
\endhead
\hline
\endfoot
\endlastfoot
\multicolumn{1}{c}{\textbf{Feature Name}} &
  \multicolumn{1}{c}{\textbf{Scale Type}} &
  \multicolumn{1}{c}{\textbf{A}} &
  \multicolumn{1}{c}{\textbf{B}} &
  \multicolumn{1}{c}{\textbf{C}} &
  \multicolumn{1}{c}{\textbf{A}} &
  \multicolumn{1}{c}{\textbf{B}} &
  \multicolumn{1}{c}{\textbf{C}} \\ \hline
revision\_count.normalized\_per\_10 &
  continuous &
  \cellcolor[HTML]{C6EFCE}{\color[HTML]{006100} 0.01} &
  \cellcolor[HTML]{C6EFCE}{\color[HTML]{006100} 0.03} &
  \cellcolor[HTML]{C6EFCE}{\color[HTML]{006100} 0.03} &
   &
   &
   \\
revision\_count.revision\_count &
  continuous &
  \cellcolor[HTML]{C6EFCE}{\color[HTML]{006100} 0.00} &
  \cellcolor[HTML]{C6EFCE}{\color[HTML]{006100} 0.05} &
  \cellcolor[HTML]{C6EFCE}{\color[HTML]{006100} 0.04} &
   &
   &
   \\
overall\_appropriateness.submetric\_scores.semantic\_similarity &
  continuous &
  \cellcolor[HTML]{FFC7CE}{\color[HTML]{9C0006} 0.30} &
  \cellcolor[HTML]{FFC7CE}{\color[HTML]{9C0006} 0.40} &
  \cellcolor[HTML]{FABF8F}{\color[HTML]{974706} 0.20} &
   &
   &
   \\
semantic\_similarity.similarity\_score &
  continuous &
  \cellcolor[HTML]{FFC7CE}{\color[HTML]{9C0006} 0.30} &
  \cellcolor[HTML]{FFC7CE}{\color[HTML]{9C0006} 0.40} &
  \cellcolor[HTML]{FABF8F}{\color[HTML]{974706} 0.20} &
   &
   &
   \\
overall\_appropriateness.appropriateness\_score &
  continuous &
  \cellcolor[HTML]{FFC7CE}{\color[HTML]{9C0006} 0.35} &
  \cellcolor[HTML]{FFC7CE}{\color[HTML]{9C0006} 0.40} &
  \cellcolor[HTML]{FABF8F}{\color[HTML]{974706} 0.14} &
   &
   &
   \\
overall\_appropriateness.submetric\_scores.thematic\_focus &
  continuous &
  \cellcolor[HTML]{FFC7CE}{\color[HTML]{9C0006} 0.42} &
  \cellcolor[HTML]{FFC7CE}{\color[HTML]{9C0006} 0.43} &
  \cellcolor[HTML]{FABF8F}{\color[HTML]{974706} 0.11} &
   &
   &
   \\
\hline
\end{longtable}